\pdfoutput=1

\documentclass[11pt]{article}

\usepackage[final]{acl}

\usepackage{times}
\usepackage{latexsym}

\usepackage[T1]{fontenc}

\usepackage[utf8]{inputenc}

\usepackage{microtype}
\usepackage{inconsolata}
\usepackage{graphicx}
\usepackage{microtype}
\usepackage{amsmath}
\usepackage{graphicx}
\usepackage{subfigure}
\usepackage{amsfonts}
\usepackage{multirow}
\usepackage{listings}
\usepackage{url}
\usepackage{CJKutf8}
\usepackage{graphicx}
\usepackage{subfigure}
\usepackage{supertabular}
\usepackage{subfigure}
\usepackage{dcolumn,booktabs}
\usepackage{tikz}
\usepackage{xspace}
\usepackage{authblk}
\usepackage[most]{tcolorbox}

\def\secref#1{\S\ref{sec:#1}}
\def\seclabel#1{\label{sec:#1}}

\newcounter{notecounter}
\newcommand{\enotesoff}{\long\gdef\enote##1##2{}}
\newcommand{\enoteson}{\long\gdef\enote##1##2{{
\stepcounter{notecounter}
{\large\bf
\hspace{1cm}\arabic{notecounter} $<<<$ ##1: ##2
$>>>$\hspace{1cm}}}}}
\enoteson
\enotesoff

\title{Understanding In-Context Machine Translation for Low-Resource Languages: A Case Study on Manchu}

\author[1,*]{\bf Renhao Pei}
\author[1,2,*]{\bf Yihong Liu}
\author[1,2]{\bf Peiqin Lin}
\author[3,$\dag$]{\bf François Yvon}
\author[1,2,$\dag$]{\bf Hinrich Sch\"utze}

\affil[]{Center for Information and Language Processing, LMU Munich \protect\\ $^{2}$Munich Center for Machine Learning (MCML) \protect\\ $^{3}$Sorbonne Université, CNRS, ISIR, France
 \protect\\ \texttt{renhaopei@gmail.com} \ \ \ \ \ \ \ \texttt{yihong@cis.lmu.de}}

\begin{document}
\maketitle

\def\thefootnote{*}\footnotetext{Equal contribution.}\def\thefootnote{\arabic{footnote}}

\def\thefootnote{$\dag$}\footnotetext{Equal advising.}\def\thefootnote{\arabic{footnote}}

\begin{abstract}
In-context machine translation (MT) with large language models (LLMs) is a promising approach for low-resource MT, as it can readily take advantage of linguistic resources such as grammar books and dictionaries.
Such resources are usually selectively integrated into the prompt so that LLMs can directly perform translation without any specific training, via their in-context learning capability (ICL).
However, the relative importance of each type of resource, e.g., dictionary, grammar book, and retrieved parallel examples, is not entirely clear.
To address this gap, this study systematically investigates how each resource and its quality affect the translation performance, with the \textbf{Manchu} language as our case study. 
To remove any prior knowledge of Manchu encoded in the LLM parameters and single out the effect of ICL, we also experiment with an enciphered version of Manchu texts.
Our results indicate that high-quality dictionaries and good parallel examples are very helpful, while grammars hardly help.
In a follow-up study, we showcase a promising application of in-context MT: parallel data augmentation as a way to bootstrap a conventional MT model. 
When monolingual data abound, generating synthetic parallel data through in-context MT offers a pathway to mitigate data scarcity and build effective and efficient low-resource neural MT systems.\footnote{We make our code and data publicly available at: \url{https://github.com/cisnlp/manchu-in-context-mt}.}
\end{abstract}

\section{Introduction}

Neural machine translation (NMT) systems have achieved remarkable performance in high-resource language pairs for which parallel sentence-level or document-level data are abundant \citep{Bahdanau2015attention,Vaswani2017transformer,tiedemann-scherrer-2017-neural,laubli-etal-2018-machine}. 
However, parallel data is scarce or even unavailable for many low-resource or endangered languages \citep{haddow-etal-2022-survey}, which prevents the training of dedicated MT systems for these languages. While multilingual models partly mitigate this issue \citep{costa-jussa-etal-2024-scaling}, they only cover a small fraction of the world's languages and their performance remains unsatisfactory for many language pairs.

On the other hand, owing to the work of field linguists,
grammatical descriptions or dictionaries are available for more than 60\% of the world's languages \citep{nordhoff2011glottolog,zhang-etal-2024-hire}.\footnote{This includes long grammatical books (24\%), short grammatical books (13\%), and grammatical sketches (25\%), according to \url{https://glottolog.org/langdoc/status}.}
Some low-resource languages are well-documented, with rich linguistic resources gathered over decades of meticulous fieldwork and analysis by linguists: this is for instance the case of Japhug, a minority Sino-Tibetan language, for which a comprehensive grammar, including plentiful glossed and translated examples, has been released by \citet{Jacques2021}.
The situation of Manchu is even more favorable, as multiple grammar books, dictionaries, and textbooks are readily available. 
Yet, all these languages are still considered low-resource in the context of data-driven MT, simply due to the scarcity of parallel data. A natural question is then to explore whether such linguistic knowledge can make up for the lack of parallel data, and help develop MT systems.

The recent emergence of LLMs seems to offer new promising ways to address this question, based on their 
in-context learning ability \citep{tanzer2024a,zhang-etal-2024-hire,hus-anastasopoulos-2024-back,merx-etal-2024-low}.
In these studies, linguistic resources such as dictionaries, parallel examples, and grammar books are integrated into the prompt and encoded together with the sentence to be translated.
We continue this line of work, trying to better analyze the role and impact of each type of linguistic knowledge that can be put to use in LLM-based machine translation systems.

For this, we perform a systematic investigation of how each component affects the in-context MT performance, with the translation from \textbf{Manchu} into English\footnote{Translation into from Manchu into Chinese is also considered in Appendix~\ref{sec:manchu2chinese}.} as a case study.
Specifically, we leverage a wide range of state-of-the-art open-source and closed-source LLMs and consider the following linguistic resources (components): dictionaries, parallel examples, grammar books, and Chain-of-Thought (CoT) prompting.
For each component, we consider several variants that vary in the amount of information or the degree of relevance to the sentence to be translated.
To quantify the influence of prior knowledge of Manchu in LLMs, we perform a character-level encipherment to disentangle the effect of LLMs’ prior knowledge of Manchu from their in-context learning ability.
In addition, we demonstrate a use case of our in-context MT system, using it as a data-augmentation tool to turn a monolingual Manchu corpus into a parallel corpus. With these synthetic parallel data incorporated into the training set, we fine-tune the mT5 model \citep{xue-etal-2021-mt5}, achieving a substantial performance gain compared to the baseline that only uses actual parallel data.

The main contributions of this work are as follows: 
\textbf{(i)} We conduct a comprehensive investigation of in-context MT for Manchu, exploring the most important knowledge sources provided in the context, highlighting the positive role of high-quality dictionaries and closely related parallel examples. 
\textbf{(ii)} Using an enciphered version of Manchu, we isolate the limited prior knowledge of Manchu encoded in the LLMs considered in our work and show that most of their translation performance depends on their in-context learning abilities.
\textbf{(iii)} We use in-context MT to generate synthetic parallel data from monolingual data of Manchu and measure how much this form of data augmentation actually benefits low-resource NMT.

\section{Related Work}

\paragraph{Low-resource NMT}
The challenges posed by limited parallel data has motivated extensive research on innovative strategies for low-resource NMT
\citep{haddow-etal-2022-survey,yazar2023lowresource}. 
Various approaches have been proposed to improve translation quality in such settings. Data augmentation techniques, such as back-translation \citep{sennrich-etal-2016-improving,edunov-etal-2018-understanding} and forward-translation \citep{bogoychev2020domaintranslationesenoisesynthetic}, have been widely used to generate synthetic parallel data and improve model performance. 
Data augmentation, coupled with unsupervised and semi-supervised methods for bilingual dictionary induction, has enabled translation with minimal parallel data, relying instead on monolingual resources \citep{Lample2018unsupervised,Artetxe2018Unsupervised}.
Transfer learning has also proven effective, where models pretrained on high-resource language pairs can be adapted to low-resource languages \citep{zoph-etal-2016-transfer, tars-etal-2022-teaching, her-kruschwitz-2024-investigating}. 
Recent advancements in multilingual NMT also show that models trained on multiple language pairs can better deal with low-resource languages \citep{ko-etal-2021-adapting,mohammadshahi-etal-2022-small,costa-jussa-etal-2024-scaling}.
Despite these advancements, achieving high-quality translation in low-resource scenarios remains a significant challenge.

\paragraph{LLM-based In-context MT for Low-Resource Languages}

Although not explicitly trained for machine translation, LLMs can perform translation by following instructions and demonstrations in the prompt \citep{NEURIPS2020_1457c0d6,lin-etal-2022-shot,vilar-etal-2023-prompting}. LLM-based MT, however, struggles with rare words that appear infrequently in the training data \citep{ghazvininejad2023dictionarybasedphraselevelpromptinglarge}. This issue is particularly pronounced for low-resource languages that are underrepresented in the LLM's training corpora \citep{workshop2023bloom176bparameteropenaccessmultilingual,touvron2023llama2openfoundation}.
To mitigate this, some studies incorporate linguistic resources into prompts, such as \textbf{dictionary entries} and \textbf{parallel sentence examples} \citep{ghazvininejad2023dictionarybasedphraselevelpromptinglarge,zhang-etal-2024-teaching}, as well as \textbf{grammars} \citep{tanzer2024a,hus-anastasopoulos-2024-back}. Some works also include \textbf{morphological analyzers} to decompose input sentences into morphemes \citep{zhang-etal-2024-hire}.
Additionally, prompting strategies such as \textbf{CoT} reasoning have been explored in the context of MT \citep{elsner-needle-2023-translating}.
However, little attention has been given to how the quality of each component affects the LLM-based in-context MT. 
Moreover, there is a lack of clear ablation studies disentangling the effects of an LLM's prior knowledge of the language and the linguistic information provided in context.
Addressing these limitations of previous studies, our work systematically investigates the role of each of these components in LLM-based MT for low-resource languages, using Manchu as a case study.

\section{Language, Data and General Setup}\seclabel{setup}

\paragraph{Manchu Language} Manchu (ISO 639-3: \texttt{mnc}) is a critically endangered Tungusic language native to Northeast China. It is the traditional language of the Manchu people and was one of the official languages of the Qing dynasty (1644-1911) of China. Because of its significant historical importance, Manchu has been extensively studied, and there exist abundant linguistic resources, including dictionaries, grammar books, and some bilingual parallel sentences, which make Manchu well-suited for our case study. A more detailed description of the Manchu language is given in Appendix~\ref{sec:appendix A}.

\paragraph{Dictionary}
We use the comprehensive dictionary from \citet{norman2020comprehensive},\footnote{ \url{https://buleku.org/home}.
}
which contains rich information such as the multiple senses for polysemous words as well as frequent collocations. It serves as our main Manchu-English lexicon.
Additionally, we compile a dictionary for Manchu suffixes based on \citep{clark1980manchu},
which contains brief explanations for each suffix.\footnote{Manchu exclusively uses suffixation, therefore neither prefixation nor circumfixation is involved.}

\paragraph{Parallel Corpus}\seclabel{parallelcorpus}
The main source of parallel data is a Manchu-Chinese dictionary \citep{1994}, which contains parallel example sentences for many dictionary entries.\footnote{Data is available from \url{https://gerel.net/}.}
We extract parallel sentences from the dictionary, followed by data-cleaning and filtering steps, to ensure that the Chinese sentences are in modern Standard Chinese.
The result is a Manchu-Chinese parallel corpus consisting of 3,520 sentence pairs, encompassing diverse genres, including everyday conversations, historical records, and literary works.
We then use the Google Cloud Translation API
to translate the Chinese sentences into English, thereby creating a Manchu-English parallel corpus.\footnote{\url{https://cloud.google.com/translate?hl=en}}
\paragraph{Monolingual Corpus}\seclabel{monolingualcorpus}
We also compile a monolingual Manchu corpus consisting of 
42,240
sentences collected from websites, encompassing a diverse range of genres.\footnote{\url{https://manc.hu/} and \url{https://gerel.net/}} During our data augmentation experiment presented in \secref{augmentation}, this monolingual Manchu corpus serves to build a synthetic Manchu-English parallel corpus.

\paragraph{Grammar}\seclabel{grammarbook}
We use two grammar books: a concise grammar book \citep{norman1965grammatical}
and a more detailed grammar from \citep{gorelova2002manchu}.


\paragraph{Evaluation Set}\seclabel{evaluationset}
We compile a test set of 337 Manchu-English parallel sentences for evaluation. 
This test set consists of 70 sentences from \textit{Nogeoldae}, a book containing dialogues in Manchu, paired with English translations \citep{zhang-etal-2024-hire}, and 267~sentence pairs extracted from \citep{di2007diary}.\footnote{\url{https://github.com/ulingga/Manchu-English_babyMT}.}
We have made sure that the parallel corpus and the evaluation set do not overlap.

\paragraph{Models}
We conduct our experiments with multiple LLMs: GPT-4o \citep{openai2024gpt4technicalreport}, DeepSeek-V3 \citep{deepseekai2024deepseekv3technicalreport}, and 
Llama3 models \citep{grattafiori2024llama3herdmodels}. For the Llama3 family, we test models of varying sizes -- 1B, 3B, 8B, and 70B -- to evaluate how model size impacts performance.

\paragraph{Evaluation Metrics}
We use BLEU \citep{papineni-etal-2002-bleu} and chrF \citep{popovic-2015-chrf} to measure the translation quality, as implemented by SacreBLEU \citep{post-2018-call}.\footnote{Signature:
\texttt{nrefs:1|case:lc|eff:no|tok:13a|smooth:\newline exp|version:2.4.3.}} Additionally, we use SBERT \citep{reimers-2019-sentence-bert}, an encoding based-metric, which assesses the semantic relatedness between a hypothesis and a reference using the cosine similarity of their embeddings (scores are multiplied by 100 to ensure a uniform magnitude).
\enote{a}{Check signature R:updated}

\section{Assessing Each Component}\seclabel{component}
  
Following the standard pipeline for in-context MT \citep{tanzer2024a,hus-anastasopoulos-2024-back,zhang-etal-2024-hire,zhang-etal-2024-teaching}, our goal is to conduct a rigorous investigation of the importance of each type of linguistic resource (component) and its quality to the translation performance. For each input Manchu sentence, a structured prompt is constructed by integrating various components. This prompt is then fed to the LLM to generate a response, from which the translation is extracted. The translation is finally evaluated against the ground truth reference using various metrics.

\paragraph{Formulation of Prompts} We represent the prompt formulation as $\pi(\cdot)$ which takes several arguments as input. Let $\mathbf{x}$ be the Manchu sentence to be translated. The simplest prompt is $\pi(\mathbf{x})$, which asks the LLM to directly translate $\mathbf{x}$ into the target language, without providing any additional information. The prompt template can be augmented by adding optional arguments as follows -- each representing one component.\footnote{Prompt templates are illustrated in Appendix~\ref{sec:appendix prompts}.}

\begin{itemize}
    \item A morphological analyzer $\mu(\cdot)$, which transforms $\mathbf{x}$ into segmented and analyzed morphemes. The result is represented as $\mu(\mathbf{x})$.
    \item Dictionary entries $\mathrm{D}$ retrieved from a bilingual dictionary $\mathcal{D}$.
    \item Parallel examples $\mathrm{P}$ retrieved from a parallel corpus $\mathcal{P}$.
    \item Grammar excerpts $\mathrm{G}$ retrieved from a grammar book $\mathcal{G}$.
    \item CoT prompting instructions $\mathrm{C}$ selected from a set of prompting varieties $\mathcal{C}$.
\end{itemize}

\begin{figure}
    \centering
    \setlength{\abovecaptionskip}{-0.2cm}
    \includegraphics[width=0.48\textwidth]{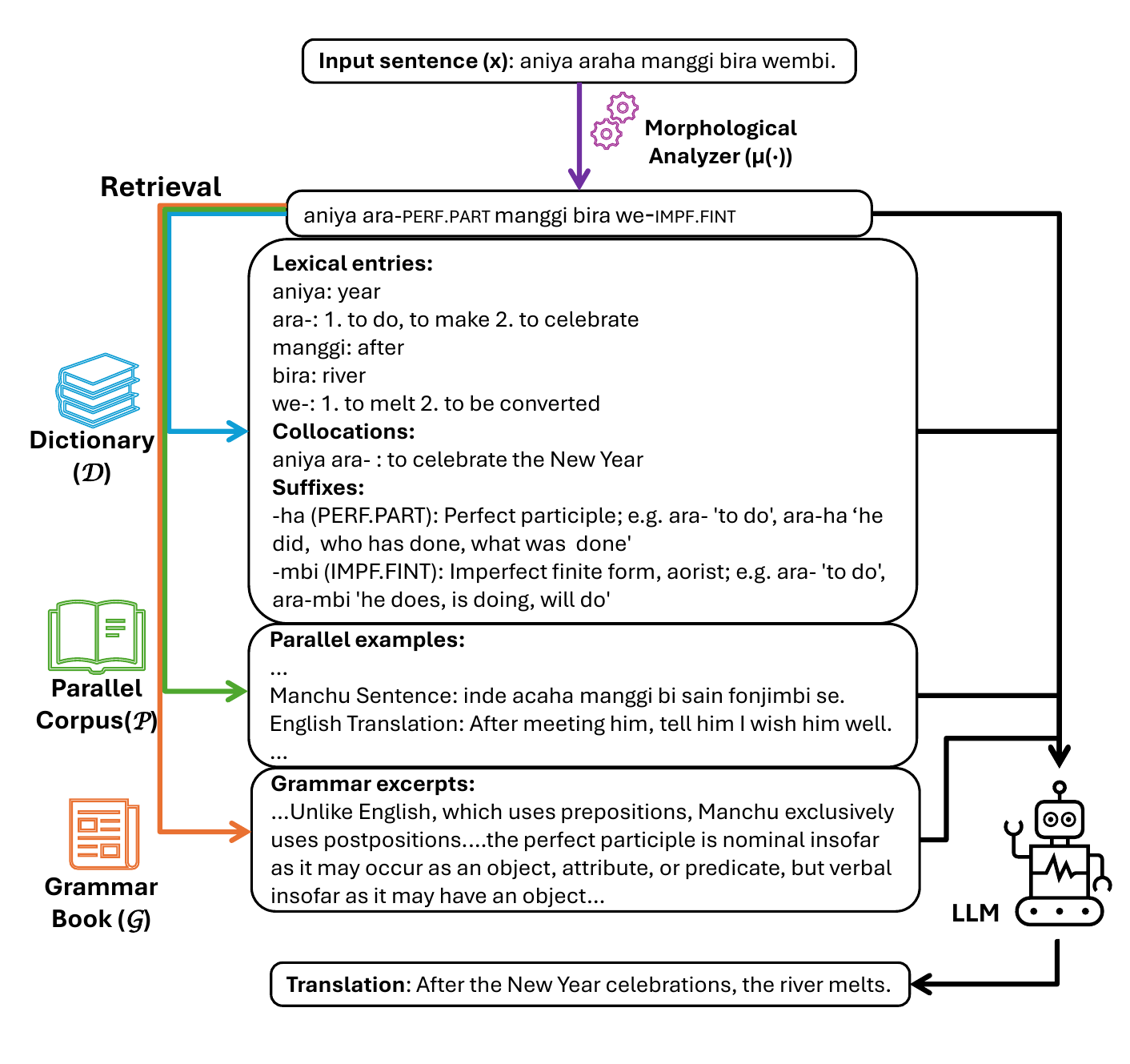}
    \caption{Illustration of the in-context MT pipeline with
    the components of $\pi$,
    i.e., $\mu(\mathbf{x})$, $\mathrm{D}$, $\mathrm{P}$ and $\mathrm{G}$.}
    \label{fig:pipeline}
\end{figure}

\paragraph{Sequential Integration of Components}\seclabel{Sequential} 
Given that many components in our pipeline have multiple implementations of varying quality,
exhaustively evaluating all possible combinations would be computationally infeasible.
Therefore, we add components to $\pi(\cdot)$ sequentially and compare performance between implementations for that component. 
The best-performing one is used as a new baseline when we evaluate the next component.
Specifically, starting with the simple baseline $\pi(\mathbf{x})$, we first add the morphological analyzer, a fundamental element for subsequent retrieval components, resulting in $\pi(\mu(\mathbf{x}))$.\footnote{We only consider one version of $\mu(\mathbf{x})$, which constitutes an essential component for all other linguistic resources.}
We then consider components that have multiple variants.
To begin with, we consider various ways to specify $\mathrm{D}$ and select the best one $\pi(\mu(\mathbf{x}), \mathrm{D}^*)$ which is the new baseline for subsequent add-ons. 
We then follow the order $\mathrm{P}$, $\mathrm{G}$, $\mathrm{C}$, resulting in
$\pi(\mu(\mathbf{x})^*, \mathrm{D}^*, \mathrm{P})$ (assessing multiple ways to select parallel examples), 
$\pi(\mu(\mathbf{x})^*, \mathrm{D}^*, \mathrm{P}^*, \mathrm{G})$ (assessing a variety of grammar excerpts), and 
$\pi(\mu(\mathbf{x})^*, \mathrm{D}^*, \mathrm{P}^*, \mathrm{G}^*, \mathrm{C})$ (assessing variants of CoT instruction).
This order prioritizes components that are expected to be most beneficial, with less helpful components introduced at later stages, as suggested by previous works \citep{zhang-etal-2024-hire,hus-anastasopoulos-2024-back}.
The pipeline is depicted in Figure~\ref{fig:pipeline}.
In the following sections, we study each component in detail and report experimental results obtained using the \textbf{GPT-4o model} on the evaluation set of 337~Manchu-English parallel sentences.

\subsection{Morphological Analysis}\seclabel{morphological_analyzer}

Morphological analysis is usually performed in a naive way in previous studies. For instance, \citet{zhang-etal-2024-hire} simply perform a dictionary look-up -- searching for \textbf{inflected word forms} in a dictionary, the coverage of which is limited.\footnote{The dictionary is from \citet{norman2020comprehensive}. It can be accessed at \url{https://buleku.org/home}.} 
As Manchu is an agglutinative language and exclusively uses suffixation, identifying word stems and suffixes is straightforward. Therefore, we implement a rule-based morphological analyzer that splits an input word into a stem and a sequence of suffixes. Both the \textbf{list of word stems} and the \textbf{set of allowed suffixes} are obtained from the dictionary. Our morphological analyzer then attempts to recursively detach a suffix from the end of a string until the remaining segment matches a known word stem. After the morphological analysis, a Manchu sentence is transformed into a list of morphemes (containing word stems and suffixes), which serves as the basis for retrieving dictionary entries, parallel examples, and grammar excerpts.

It is possible for a Manchu word to have multiple analyses. For example, \textit{tere} could be a demonstrative pronoun meaning ``that'', or it could be analyzed as \textit{te-re}, meaning ``sitting'' as the present participle of the verb \textit{te} ``to sit''. 
In such cases, we include all possible analyses in the prompt and let the LLM resolve the ambiguity by selecting the most contextually appropriate interpretation, as shown in Table~\ref{example:disambiguation} of Appendix~\ref{sec:output_examples}.

\subsection{Dictionary}
The dictionary $\mathcal{D}$ comprises lexical entries, suffixes, and their corresponding collocations. These three elements -- lexical entries, suffixes, and collocations -- form the foundation of our three variants:
\enote{a}{Could we use 'glosses' instead of 'explanations'? R: I think 'explanations' is better here, as our entry includes more detailed multi-word explanations than simple gloss}

\begin{itemize} 
    \item $\mathrm{D^l}$ includes only the \textbf{lexical} entries retrieved for each word in the input sentence, without explanation\footnote{Explanations (in English) document the meaning and function of each suffix. See example in Appendix~\ref{sec:output_examples}.} for the suffixes.
    \item $\mathrm{D^{l+s}}$ includes both the \textbf{lexical} entries and explanations for \textbf{suffixes} for all morphemes appearing in the input sentence.
    \item $\mathrm{D^{l+s+c}}$ includes the \textbf{lexical} entries, explanations of \textbf{suffixes}, and the \textbf{collocations} for all morphemes in the input sentence.
\end{itemize}

\begin{table}[!ht]
\setlength{\abovecaptionskip}{0cm}
\setlength{\belowcaptionskip}{-0.4cm}
\setlength{\tabcolsep}{3mm}{}
\centering
\begin{tabular}{lrrr}
\hline
\textbf{Variant} & \textbf{BLEU} & \textbf{chrF} & \textbf{SBERT} \\
\hline
$\pi(\mathbf{x})$  & 3.44 & 21.86 & 34.21 \\
\hline
$\pi(\mu(\mathbf{x}))$  & 3.10 & 21.68 & 33.49 \\
w/ $\mathrm{D^{l}}$  & 7.40 & 31.84 & 58.91 \\
w/ $\mathrm{D^{l+s}}$  & 7.47 & \textbf{32.93} & 59.78 \\
w/ $\mathrm{D^{l+s+c}}$  & \textbf{7.55} & 32.71 & \textbf{61.07} \\
\hline
\end{tabular}
\caption{MT scores for direct prompting $\pi(\mathbf{x})$ and prompting with morphologically analyzed sentences $\pi(\mu(\mathbf{x}))$, and with \textbf{dictionary} entries of increasing complexities. \textbf{Bold}: best result for each column.}
\label{tab:dict}
\end{table}

\paragraph{Comprehensive dictionary entries are important.}
As shown in Table~\ref{tab:dict}, using a morphological analyzer alone is not helpful -- $\pi(\mu(\mathbf{x}))$ performs worse than $\pi(\mathbf{x})$. This is expected as simply transforming the input sentence to a segmented list of morphemes does not provide the model much knowledge about Manchu.
Once the explanations for the lexical entries are included, performance improves significantly. 
The translation quality can be further improved with the inclusion of suffixes ($\mathrm{D^{l+s}}$) and then collocations ($\mathrm{D^{l+s+c}}$).
Although the chrF score of $\mathrm{D^{l+s+c}}$ is slightly lower than $\mathrm{D^{l+s}}$, both the BLEU and SBERT scores suggest that the $\mathrm{D^{l+s+c}}$ delivers the best overall translations.
Therefore, $\pi(\mu(\mathbf{x}), \mathrm{D}^{l+s+c})$ will be used as a new baseline when assessing the following component.
We illustrate the benefits of dictionary information (lexical entries, suffixes, and collocations) for translation in Tables~\ref{example:disambiguation} and~\ref{tab:output_dict} in Appendix~\ref{sec:output_examples}.

\subsection{Parallel Examples}
Parallel examples are drawn from the corpus introduced in~\secref{parallelcorpus}.
Ideally, these parallel examples $\mathcal{P}$ should closely resemble the input sentence, as higher similarity with the source text is known to improve translation quality \citep{zhang-etal-2024-teaching}.
To explore this, we construct three variants for $\mathcal{P}$, each exhibiting a different degree of similarity:

\begin{itemize} 
\item $\mathrm{P^r}$ includes 10 parallel sentences \textbf{randomly} selected as few-shot examples.
\item $\mathrm{P^{d}}$ includes up to 10 parallel sentences retrieved based on shared terms. As the parallel sentences are extracted from the \textbf{dictionary}, they are originally meant to illustrate the meaning of a specific dictionary entry. Therefore, for a lexeme in the input sentence, the parallel examples for its dictionary entry are retrieved. 
\item $\mathrm{P^{bm}}$ includes 10 parallel sentences retrieved using the \textbf{BM25} algorithm \citep{robertson1995okapi} implemented by Rank-BM25.\footnote{\url{https://github.com/dorianbrown/rank_bm25}}
The terms used by the retriever are morphemes segmented by the analyzer of \secref{morphological_analyzer}.
\end{itemize}

\begin{table}[!ht]
\setlength{\abovecaptionskip}{0cm}
\setlength{\belowcaptionskip}{-0.4cm}
\setlength{\tabcolsep}{2mm}{}
\centering
\begin{tabular}{lrrr}
\hline
\textbf{Variant} & \textbf{BLEU} & \textbf{chrF} & \textbf{SBERT} \\
\hline
$\pi(\mu(\mathbf{x}), \mathrm{D^*})$   & 7.55 & 32.71 & 61.07 \\
w/ $\mathrm{P^{r}}$  & 7.66 & 32.94 & 60.85 \\
w/ $\mathrm{P^{d}}$  & 8.10 & 32.95 & 61.04 \\
w/ $\mathrm{P^{bm}}$  & \textbf{8.84} & \textbf{33.72} & \textbf{61.35} \\
\hline
\end{tabular}
\caption{Performance comparison between the baseline (no parallel examples) and 3 ways to select \textbf{parallel examples}. \textbf{Bold}: best result for each column.}
\label{tab:para}
\end{table}

\paragraph{More similar parallel examples improve the translation.}
As shown in Table~\ref{tab:para},  randomly retrieved parallel examples provide only a slight improvement over the baseline, as they do not seem to introduce much useful information into the context. On the other hand, selecting parallel examples that are similar to the input sentence yields more noticeable improvements (see lines $\mathrm{P^{d}}$ and $\mathrm{P^{bm}}$ in Table~\ref{tab:para}).
$\mathrm{P^{bm}}$ achieves the best performance across all 3 evaluation metrics, as BM25 aims to retrieve parallel examples that are globally similar to the input sentence; $\pi(\mu(\mathbf{x}), \mathrm{D}^{l+s+c}, \mathrm{P^{bm}})$ will be used as a new baseline when assessing the following component.
An example of how parallel examples help translation is in Table~\ref{example:output_para} in Appendix~\ref{sec:output_examples}.

\subsection{Grammar}\seclabel{Grammar}
As mentioned in
\secref{grammarbook}, two grammar books -- a short and a more detailed one -- serve as source materials.
For each book, we manually compile 26 tuples consisting of (\textit{feature, excerpt}), in which each Manchu grammatical feature is paired with the corresponding excerpt from the short or the long grammar book.
With our morphological analyzer, we extract a set of grammatical features from the source Manchu sentence and generate a tailored grammar combination accordingly, consisting of only excerpts that are relevant to that sentence. 
This approach is much more efficient than dumping the entire grammar book into the context.
We consider 3 ways to retrieve excerpts $\mathrm{G}$ from grammar books:
\begin{itemize} 
\item $\mathrm{G^{s}}$ is a combination of grammar excerpts, retrieved from the \textbf{short} book.
\item $\mathrm{G^l}$ is a combination of grammar excerpts, retrieved from the \textbf{long} grammar book with more detailed explanations. 
    \item $\mathrm{G^{l+p}}$ additionally adds \textbf{parallel} examples that illustrate the grammar excerpts, which are originally included in the \textbf{long} grammar book.
\end{itemize}
In addition to the excerpts, we include a fixed paragraph shared by all variants, which contains basic information about the word order and typological features of Manchu
(see Appendix~\ref{sec:appendix prompts}).

\begin{table}[ht!]
\setlength{\tabcolsep}{2mm}{}
\centering
\begin{tabular}{lrrr}
\hline
\textbf{Variant} & \textbf{BLEU} & \textbf{chrF} & \textbf{SBERT} \\
\hline
$\pi(\mu(\mathbf{x}),\mathrm{D^{*}},\mathrm{P^{*}})$ & 8.84 & 33.72 & \textbf{61.35} \\
w/ $\mathrm{G^{s}}$  & 8.26 & 33.12 & 60.70 \\
w/ $\mathrm{G^{l}}$  & 8.46 & \textbf{33.79} & 61.17 \\
w/ $\mathrm{G^{l+p}}$  & \textbf{8.90} & 33.77 & 60.40 \\
\hline
\end{tabular}
\caption{Performance comparison between the baseline without \textbf{grammar} and 3 different variants of retrieving grammar excerpts. \textbf{Bold}: best result for each column.}
\label{tab:grammar}
\end{table}

\paragraph{Grammars hardly help.}
As shown in Table \ref{tab:grammar}, $\mathrm{G^{s}}$ yields scores worse than the baseline. With more detailed explanations, $\mathrm{G^{l}}$ leads to a slight improvement in chrF score, and when further accompanied by parallel examples, $\mathrm{G^{l+p}}$ leads to a small improvement in BLEU score. 
Nevertheless, compared to the performance reported for the other components, i.e., dictionary and parallel examples (cf. Tables~\ref{tab:dict} and \ref{tab:para}), the improvement seems marginal and is not reflected in SBERT scores.
This suggests that grammars do not help much in in-context MT, 
which is consistent with the findings reported by \citet{aycock2024llmsreallylearntranslate}.
Nevertheless, we have found instances where grammar explanations could aid translation, such as the example of Table \ref{example:grammar} in Appendix~\ref{sec:output_examples}.
Moreover, since the next component -- CoT -- involves grammatical annotation and syntactic analysis, which are closely tied to the information provided in the grammar excerpts, we will still include the grammar component in the new baseline for assessing the CoT component. The variant $\mathrm{G^{l+p}}$ is selected based on the BLEU score.

\subsection{Chain-of-Thought}

CoT prompting instructs LLMs to generate a series of intermediate results
before solving the final task \citep{cot2022wei}. We draw CoT prompt templates from \texttt{LingoLLM} \citep{zhang-etal-2024-hire}, which are explicit instructions provided in the context. 
We consider 2 variants for CoT prompting $\mathrm{C}$:
\begin{itemize} 
\item $\mathrm{C^a}$ asks the LLM to \textbf{annotate} the grammatical and semantic features of each word in the sentence before computing the translation. 
\item $\mathrm{C^{a+s}}$ asks the LLM to proceed step by step, first to annotate the grammatical and semantic features of each word, then analyze the sentence's \textbf{syntactic structure}, and finally produce the translation. 
\end{itemize}

\begin{table}[ht!]
\setlength{\tabcolsep}{1.2mm}{}
\centering
\begin{tabular}{lccc}
\hline
\textbf{Variant} & \textbf{BLEU} & \textbf{chrF} & \textbf{SBERT} \\
\hline
$\pi(\mu(\mathbf{x}),\mathrm{D^{*}},\mathrm{P^{*}},\mathrm{G^{*}})$ & \textbf{8.90} & \textbf{33.77} & \textbf{60.40} \\
w/ $\mathrm{C^{a}}$  & 8.01 & 33.13 & 59.81 \\
w/ $\mathrm{C^{a+s}}$  & 8.49 & 33.43 & 59.01 \\
\hline
\end{tabular}
\caption{Performance comparison between the baseline without CoT prompting and 2 variants of CoT. \textbf{Bold}: best result for each column.}
\label{tab:cot}
\end{table}

\paragraph{CoT does not help the model generate better translations.}
Explicitly prompting the model to perform intermediate generation steps results in a noticeable decline in both $\mathrm{C^{a}}$ and $\mathrm{C^{a+s}}$. This aligns with the findings of \citet{elsner-needle-2023-translating},
where CoT does not improve performance.
This discrepancy seems to arise from erroneous or incomplete deductions within the intermediate steps (cf. Table \ref{example:cot_unhelpful} in Appendix~\ref{sec:output_examples}). This further indicates that, even with the CoT prompting, the model is still unable to effectively utilize the grammar. Consequently, we exclude the CoT component and the Grammar component from our final pipeline.

\section{In-Depth Analysis of Performance} \seclabel{analysis}

\subsection{Performance Across Models}

We have so far used the GPT-4o model to assess the importance of each component and its quality, finding that the best setting is  $\pi(\mu(\mathbf{x}),\mathrm{D^{l+s+c}},\mathrm{P^{bm}})$.
We now study performance variation across models for this setting.
Results are in Table~\ref{tab:model_size}.

\paragraph{Model size matters.}
The smallest model, i.e., Llama3-1B, yields an extremely low BLEU score of 0.27. When manually checking the translation, we found that the Llama3-1B model often does not follow the instructions, generating outputs where the translation is difficult to extract or missing entirely. With the size increase in the Llama3 family, we see a consistent improvement in translation scores. Through manual inspection of the translations from varying model sizes (see Table~\ref{example:size} in Appendix~\ref{sec:output_examples}), we observe that larger models not only exhibit better instruction-following abilities but are also better at leveraging the information included in the context. Therefore, we hypothesize that LLM-based MT relies on both good instruction-following and in-context learning abilities, which are closely related to the model size. 

\begin{table}
\centering
\begin{tabular}{lrrr}
\hline
\textbf{Model} & \textbf{BLEU} & \textbf{chrF} & \textbf{SBERT} \\
\hline
Llama3-1B & 0.27 & 9.95 & 16.37 \\
Llama3-3B  & 1.81 & 21.95 & 38.46 \\
Llama3-8B  & 3.05 & 26.59 & 49.10 \\
Llama3-70B  & 6.31 & 31.01 & 56.82 \\
\hline
GPT-4o  & 8.84 & 33.72 & 61.35 \\
\hline
DeepSeek-V3 & \textbf{12.35} & \textbf{37.93} & \textbf{65.64} \\
\hline
\end{tabular}
\caption{Performance of various LLMs using the best setting, i.e., $\pi(\mu(\mathbf{x}),\mathrm{D^{l+s+c}},\mathrm{P^{bm}})$. \textbf{Bold}: best result for each column.}%
\label{tab:model_size}
\end{table}

\paragraph{The performance could be underestimated}\label{performance_underestimated}
The best performance is obtained with 
DeepSeek-V3, achieving BLEU scores of 12.35. The score is still low, especially when compared with LLM-based translation for high-resource languages \citep{alves-etal-2023-steering,sia-etal-2024-anti}.
However, we often observe that the in-context translations are semantically close to the reference, yet exhibit significant differences in wordings, suggesting that BLEU and chrF scores actually underestimate the MT quality, as illustrated by the example in Table~\ref{tab:sbert_scores}.
When assessed with SBERT, the best-performing model (DeepSeek-V3) achieves a score of 65.64, indicating a strong semantic similarity between the translation and the reference.

\begin{table}[h!]
\centering
\begin{tabular}{p{1.9cm}p{5cm}}
\hline
Input: & \textit{ereci julesi gurgu elgiyen} \\
\hline
Translation: & From this point forward, wild animals are abundant. \\
Reference: & From there onwards beasts were plentiful. \\
\hline
\textbf{BLEU:} 4.99 & \textbf{chrF:} 24.02 \,\,\,\, \textbf{SBERT:} 62.42\\
\end{tabular}
\caption{An example where BLEU and chrF scores (\textbf{sentence-level}) underestimate the translation quality, while SBERT better reflects the translation quality.}
\label{tab:sbert_scores}
\end{table}

\subsection{Exposing Prior Knowledge of Manchu with Character-Substitution Cipher} \seclabel{encryption}

\begin{figure}
    \centering
    \includegraphics[width=0.48\textwidth]{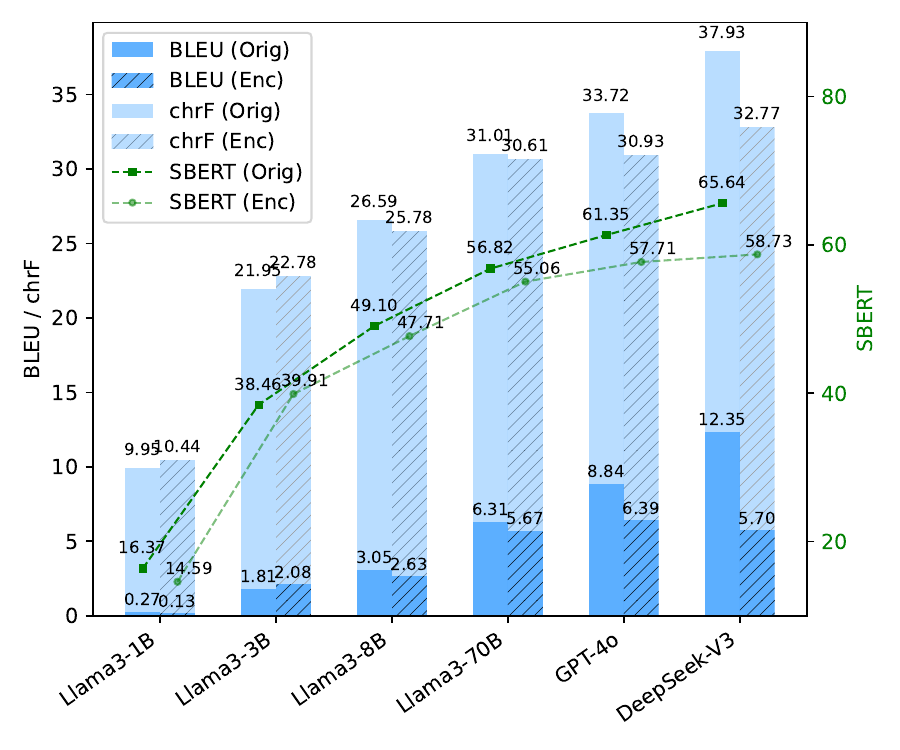}
    \caption{Performance comparison between enciphered $\pi(\mu(\mathbf{x})_{e}, \mathrm{D^{l+s+c}_{e}}, \mathrm{P^{bm}_{e}})$ and original $\pi(\mu(\mathbf{x})), \mathrm{D^{l+s+c}}, \mathrm{P^{bm}})$ across multiple LLMs.}
    \label{fig:plot2}
\end{figure}

We have so far assumed that the MT performance of LLMs is mostly attributed to their \textbf{in-context abilities}, rather than to some \textbf{prior knowledge of Manchu} that can possibly be acquired during its training stage. 
To explore this question, we create a ``fake Manchu'' aimed at eliminating this possible confounding factor. As \citet{yuan2024gpt4smartsafestealthy} and \citet{marmonier2025explicitlearningllmmachine} have demonstrated, LLMs' prior knowledge can be bypassed using substitution ciphers.
In this work, we ``encipher'' all Manchu tokens by a simple character-level substitution cipher as follows:
each vocalic character in Manchu (\textit{a, e, i, o, u}), is substituted with the \textbf{next} character in this list, e.g., $\textit{a} \rightarrow \textit{e}$, $\textit{e} \rightarrow \textit{i}$, and $\textit{u} \rightarrow \textit{a}$. The same substitution rule applies to consonantal characters in (\textit{b, c, d, f, g, h, j, k, l, m, n, p, q, r, s, t, v, w, x, y, z}). Using this scheme, a Manchu token \textit{amban} is enciphered as \textit{encep}.

The encipherment applies to all tokens in the input Manchu sentence as well as the linguistic resources involving Manchu, such as dictionary entries and parallel examples,
while the English parts remain unchanged.
This approach ensures that the LLM can only rely on the information provided in the prompt and its in-context learning ability.
Enciphered prompts are denoted with the subscript $_{e}$ as in: $\pi(\mu(\mathbf{x})_{e}, \mathrm{D_{e}}, \mathrm{P_{e}})$. We experiment with the original template $\pi(\mu(\mathbf{x}), \mathrm{D^{l+s+c}}, \mathrm{P^{bm}})$ and its enciphered version $\pi(\mu(\mathbf{x})_{e}, \mathrm{D^{l+s+c}_{e}}, \mathrm{P^{bm}_{e}})$.\footnote{We use $\pi(\mu(\mathbf{x}), \mathrm{D^{l+s+c}}, \mathrm{P^{bm}})$ because the grammar excerpts contain a mixture of Manchu and English tokens, which makes it difficult to encipher only the Manchu tokens.}
The results achieved for multiple LLMs are in Figure~\ref{fig:plot2}.

\paragraph{LLMs already know some Manchu.}
The performance of the enciphered version tends to be slightly lower than the original version for all LLMs. This suggests that all models have seen some Manchu in their pretraining stage, possibly due to contamination -- pretraining corpora often contain significant amounts of non-English texts, including many low-resourced ones \citep{blevins-zettlemoyer-2022-language}. The performance drop is particularly noticeable for DeepSeek-V3. We hypothesize that DeepSeek-V3 has seen more Manchu data during its pretraining stage because it was trained on large Chinese corpora, which may contain more Manchu texts.

\paragraph{LLMs rely more on their in-context learning ability.} 
Even though all LLMs have some prior knowledge of Manchu, as indicated by the drop in performance from the original version to the enciphered version, the enciphered versions can still achieve comparable results, and the performance gap remains relatively small (except for DeepSeek-V3).
This confirms that LLMs are not fully relying on their prior knowledge, but are rather mainly depending on their in-context learning ability. This argument can be further supported by the consistent performance improvement for both the original and the enciphered Manchu texts when increasing the model size. Since the Llama3 family models are trained on the same data, the observed performance gain, such as from Llama3-1B to Llama3-70B, should be largely attributable to the enhanced in-context learning capabilities of the larger model.

\subsection{Validating the Translation Quality Through Human Evaluation} \seclabel{human eval}
In order to further validate the quality of our MT outputs beyond automatic evaluation metrics such as BLEU, chrF, and SBERT, we have conducted a human evaluation in the form of the Direct Assessment (DA) \citep{graham-etal-2013-continuous}. 
Specifically, we have recruited 3 Manchu language experts who are fluent in both Manchu and English, and have asked them to rate how adequately the English translations express the meaning of their corresponding source sentences in Manchu, on a continuous scale from 0 to 100. 
The complete instruction given to the human raters is in Appendix~\ref{sec:appendix instructions}.

For the human evaluation, we randomly select 33 sentences from the evaluation set described in \secref{evaluationset}.
For each sentence, we include the MT outputs of 3 variants using the GPT-4o model: $\pi(\mathbf{x})$ (direct prompting), $\pi(\mu(\mathbf{x}), \mathrm{D^{l+s+c}}, \mathrm{P^{bm}})$ (the best setting), and $\pi(\mu(\mathbf{x})_{e}, \mathrm{D^{l+s+c}_{e}}, \mathrm{P^{bm}_{e}})$ (enciphered version of the best setting), resulting in a total of 99 evaluation items. 
The identities of the system variants are anonymized. 
In addition, the order of the items is randomized for the evaluation. 

To account for potential differences in how individual raters use the scoring scale, the raw DA scores are normalized to z-scores before being aggregated across raters. The inter-rater agreement among the three raters is strong, with an average Pearson correlation coefficient $r = 0.864$.

\begin{table}
\centering
\begin{tabular}{lrr}
\hline
\textbf{Variant} & \textbf{DA score} & \textbf{z-score} \\
\hline
$\pi(\mathbf{x})$ & 29.04 & -0.63 \\
$\pi(\mu(\mathbf{x})_{e}, \mathrm{D^{l+s+c}_{e}}, \mathrm{P^{bm}_{e}})$ & 56.12 & 0.19 \\
$\pi(\mu(\mathbf{x}), \mathrm{D^{l+s+c}}, \mathrm{P^{bm}})$ & \textbf{64.47} & \textbf{0.44} \\
\hline
\end{tabular}
\caption{Average DA scores and z-scores of the MT outputs across the 3 variants $\pi(\mathbf{x})$, $\pi(\mu(\mathbf{x})_{e}, \mathrm{D^{l+s+c}_{e}}, \mathrm{P^{bm}_{e}})$, and $\pi(\mu(\mathbf{x}), \mathrm{D^{l+s+c}}, \mathrm{P^{bm}})$. \textbf{Bold}: best result for each column.}
\label{tab:human_eval}
\end{table}

We report the average z-scores as well as average raw DA scores in Table~\ref{tab:human_eval}. The results show that the 2 variants enhanced with dictionary entries and parallel examples achieve substantially higher scores compared to the baseline $\pi(\mathbf{x})$. This further validates the effectiveness of our proposed pipeline. 
Moreover, $\pi(\mu(\mathbf{x}), \mathrm{D^{l+s+c}}, \mathrm{P^{bm}})$ achieves higher average scores than the enciphered version $\pi(\mu(\mathbf{x})_{e}, \mathrm{D^{l+s+c}_{e}}, \mathrm{P^{bm}_{e}})$.

We run the Wilcoxon rank-sum test to test the statistical significance. The results indicate that both $\pi(\mu(\mathbf{x})_{e}, \mathrm{D^{l+s+c}_{e}}, \mathrm{P^{bm}_{e}})$ and $\pi(\mu(\mathbf{x}), \mathrm{D^{l+s+c}}, \mathrm{P^{bm}})$ differ significantly from the baseline $\pi(\mathbf{x})$, both with $p < 0.001$. 
On the other hand, although the enciphered version has lower average scores, the difference between $\pi(\mu(\mathbf{x})_{e}, \mathrm{D^{l+s+c}_{e}}, \mathrm{P^{bm}_{e}})$ and $\pi(\mu(\mathbf{x}), \mathrm{D^{l+s+c}}, \mathrm{P^{bm}})$ is not statistically significant, with $p = 0.27$.
This also aligns with our previous finding using the automatic metrics, that the performance gap between the enciphered and the original version is relatively small: LLMs rely more on their in-context learning ability.

\section{NMT Data Augmentation}\seclabel{augmentation}

We present a follow-up study where we use our in-context MT system to generate more parallel data for training an NMT model. This data augmentation approach follows the \textit{forward-translation} method \citep{burlot-yvon-2018-using,bogoychev2020domaintranslationesenoisesynthetic}.

\paragraph{Translating Monolingual Corpus.} 
Specifically, we use our in-context MT system to translate 42,240 sentences
from the monolingual Manchu corpus (cf.~\secref{monolingualcorpus}) into English, using our best-performing method $\pi(\mu(\mathbf{x})), \mathrm{D^{l+s+c}}, \mathrm{P^{bm}})$ with DeepSeek-V3. The resulting synthetic parallel corpus is combined with the real parallel corpus to train an NMT model of Manchu-to-English.

\paragraph{Fine-Tuning mT5.}
We fine-tune mT5-small \citep{xue-etal-2021-mt5}, an encoder-decoder multilingual pre-trained model on the Manchu-to-English translation task. To systematically assess the impact of synthetic data, we use different data-mixing strategies, e.g., only real parallel data, or additionally with synthetic data that is several times larger than the real data. The performance is evaluated on the same evaluation set of 337 parallel sentences (cf. \secref{evaluationset}).

\begin{figure}
    \centering
    \includegraphics[width=0.48\textwidth]{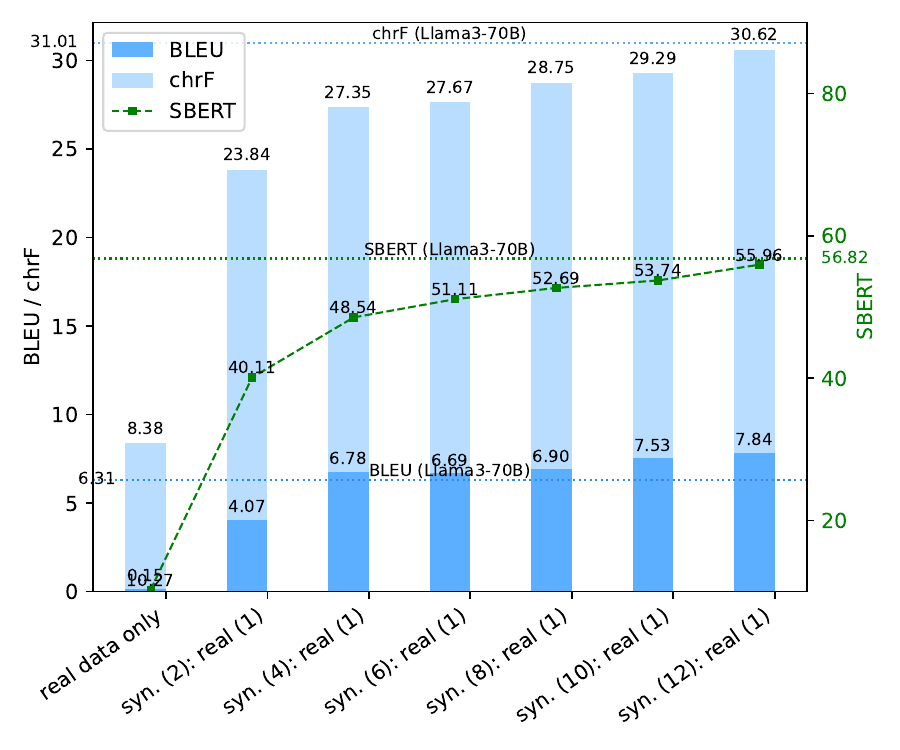}
    \caption{Performance comparison of the fine-tuned mT5 model using only real parallel data versus incorporating varying proportions of synthetic parallel data generated by our in-context MT system. We observe a steady improvement in performance as more synthetic parallel data is added, ultimately achieving scores that match the in-context MT results of Llama3-70B.}
    \label{fig:plot_finetune}
\end{figure}

Figure \ref{fig:plot_finetune} presents the results.
The model trained exclusively on real data performs extremely badly across all metrics, suggesting that 3,520 parallel sentences are insufficient for training an effective NMT model. However, as more synthetic parallel data is introduced, performance improves consistently. 
The best-performing model -- trained with real data and synthetic parallel data that are 12 times larger than the real data -- achieves results comparable to or even surpassing Llama3-70B. 
The resulting fine-tuned mT5-small model only contains around 300M parameters and is significantly more efficient than a 70B-parameter in-context MT system. 
This study underscores the potential of leveraging in-context MT for data augmentation, enabling the development of more effective and efficient NMT models for low-resource languages.

\section{Conclusion}

In this paper, we conduct a comprehensive investigation of in-context MT for low-resource languages, using Manchu as a case study. 
We examine the impact of different types of resources and the quality of each component on translation performance. 
Our findings highlight that high-quality dictionaries and properly retrieved parallel examples are the most influential factors, while grammar
and CoT prompting appears to have no noticeable benefit.
Furthermore, through the encipherment experiment, we disentangle the effects of LLMs' prior knowledge of Manchu from their in-context learning ability. 
Our results show that while LLMs possess some prior knowledge of the language, 
they primarily rely on in-context learning for translation.
Finally, our follow-up study shows a practical application of in-context MT: generating synthetic parallel data. 
This approach has the potential to enhance NMT systems, offering a viable strategy for improving translation in low-resource languages.

\section*{Limitations}
Our current work only includes a single language, Manchu, as a case study. Although our encipherment method can be considered a generalization effort applicable to any language unfamiliar to the LLMs, the encipherment did not alter the fundamental properties of Manchu, which is an agglutinative language characterized by a relatively clear separation between morphemes. It is not fully clear whether our findings extend to other typologically distinct languages.

We have only focused on the translation direction from Manchu to English and have not explored the reverse direction. 
However, if the goal is to produce synthetic parallel data of good quality, we believe it is advantageous to translate authentic low-resource language into a high-resource language that the LLM is proficient in. This ensures fluency and authenticity of the texts in both the source and target languages.

Lastly, we have only explored a limited range of CoT strategies. Our current results indicate that the extra CoT steps often introduce new errors, resulting in a poorer final translation. Future work could investigate ways to mitigate these undesired effects, such as through better prompt engineering or by providing guiding examples for the CoT process.

\section*{Acknowledgments}
This research was supported by DFG (grant SCHU 2246/14-1). François Yvon has been partly funded by the French National Funding Agency
(ANR) under the France 2030 program (ref. ANR-23-IACL-0007) and the
Tralalam Project (ref. ANR-23-IAS1-0006).
We are deeply thankful to Fresco Sam-Sin of the Manchu Foundation and Professor Hitoshi Kuribayashi from the Tohoku University, for generously granting us permission to use the digitized Manchu materials available on their websites. We also sincerely thank Manchu experts Chen Chen, Sulfa, and Zuoteng Li for their valuable contributions as raters in the human evaluation.

\bibliography{custom}

\begin{thebibliography}{56}
\providecommand{\natexlab}[1]{#1}

\bibitem[{Achiam et~al.(2024)Achiam, Adler, Agarwal, Ahmad, Akkaya, Aleman,
  Almeida, Altenschmidt, Altman, Anadkat
  et~al.}]{openai2024gpt4technicalreport}
Josh Achiam, Steven Adler, Sandhini Agarwal, Lama Ahmad, Ilge Akkaya,
  Florencia~Leoni Aleman, Diogo Almeida, Janko Altenschmidt, Sam Altman,
  Shyamal Anadkat, et~al. 2024.
\newblock \href {https://arxiv.org/abs/2303.08774} {Gpt-4 technical report}.
\newblock \emph{Preprint}, arXiv:2303.08774.

\bibitem[{Alves et~al.(2023)Alves, Guerreiro, Alves, Pombal, Rei, de~Souza,
  Colombo, and Martins}]{alves-etal-2023-steering}
Duarte Alves, Nuno Guerreiro, Jo{\~a}o Alves, Jos{\'e} Pombal, Ricardo Rei,
  Jos{\'e} de~Souza, Pierre Colombo, and Andre Martins. 2023.
\newblock \href {https://doi.org/10.18653/v1/2023.findings-emnlp.744} {Steering
  large language models for machine translation with finetuning and in-context
  learning}.
\newblock In \emph{Findings of the Association for Computational Linguistics:
  EMNLP 2023}, pages 11127--11148, Singapore. Association for Computational
  Linguistics.

\bibitem[{Artetxe et~al.(2018)Artetxe, Labaka, Agirre, and
  Cho}]{Artetxe2018Unsupervised}
Mikel Artetxe, Gorka Labaka, Eneko Agirre, and Kyunghyun Cho. 2018.
\newblock \href {https://openreview.net/forum?id=Sy2ogebAW} {Unsupervised
  neural machine translation}.
\newblock In \emph{6th International Conference on Learning Representations,
  {ICLR} 2018, Vancouver, BC, Canada, April 30 - May 3, 2018, Conference Track
  Proceedings}. OpenReview.net.

\bibitem[{Aycock et~al.(2024)Aycock, Stap, Wu, Monz, and
  Sima'an}]{aycock2024llmsreallylearntranslate}
Seth Aycock, David Stap, Di~Wu, Christof Monz, and Khalil Sima'an. 2024.
\newblock \href {https://arxiv.org/abs/2409.19151} {Can {LLMs} really learn to
  translate a low-resource language from one grammar book?}
\newblock \emph{Preprint}, arXiv:2409.19151.

\bibitem[{Bahdanau et~al.(2015)Bahdanau, Cho, and
  Bengio}]{Bahdanau2015attention}
Dzmitry Bahdanau, Kyunghyun Cho, and Yoshua Bengio. 2015.
\newblock \href {http://arxiv.org/abs/1409.0473} {Neural machine translation by
  jointly learning to align and translate}.
\newblock In \emph{3rd International Conference on Learning Representations,
  {ICLR} 2015, San Diego, CA, USA, May 7-9, 2015, Conference Track
  Proceedings}.

\bibitem[{Blevins and Zettlemoyer(2022)}]{blevins-zettlemoyer-2022-language}
Terra Blevins and Luke Zettlemoyer. 2022.
\newblock \href {https://doi.org/10.18653/v1/2022.emnlp-main.233} {Language
  contamination helps explains the cross-lingual capabilities of {E}nglish
  pretrained models}.
\newblock In \emph{Proceedings of the 2022 Conference on Empirical Methods in
  Natural Language Processing}, pages 3563--3574, Abu Dhabi, United Arab
  Emirates. Association for Computational Linguistics.

\bibitem[{Bogoychev and
  Sennrich(2020)}]{bogoychev2020domaintranslationesenoisesynthetic}
Nikolay Bogoychev and Rico Sennrich. 2020.
\newblock \href {https://arxiv.org/abs/1911.03362} {Domain, translationese and
  noise in synthetic data for neural machine translation}.
\newblock \emph{Preprint}, arXiv:1911.03362.

\bibitem[{Brown et~al.(2020)Brown, Mann, Ryder, Subbiah, Kaplan, Dhariwal,
  Neelakantan, Shyam, Sastry, Askell, Agarwal, Herbert-Voss, Krueger, Henighan,
  Child, Ramesh, Ziegler, Wu, Winter, Hesse, Chen, Sigler, Litwin, Gray, Chess,
  Clark, Berner, McCandlish, Radford, Sutskever, and
  Amodei}]{NEURIPS2020_1457c0d6}
Tom Brown, Benjamin Mann, Nick Ryder, Melanie Subbiah, Jared~D Kaplan, Prafulla
  Dhariwal, Arvind Neelakantan, Pranav Shyam, Girish Sastry, Amanda Askell,
  Sandhini Agarwal, Ariel Herbert-Voss, Gretchen Krueger, Tom Henighan, Rewon
  Child, Aditya Ramesh, Daniel Ziegler, Jeffrey Wu, Clemens Winter, Chris
  Hesse, Mark Chen, Eric Sigler, Mateusz Litwin, Scott Gray, Benjamin Chess,
  Jack Clark, Christopher Berner, Sam McCandlish, Alec Radford, Ilya Sutskever,
  and Dario Amodei. 2020.
\newblock \href
  {https://proceedings.neurips.cc/paper_files/paper/2020/file/1457c0d6bfcb4967418bfb8ac142f64a-Paper.pdf}
  {Language models are few-shot learners}.
\newblock In \emph{Advances in Neural Information Processing Systems},
  volume~33, pages 1877--1901. Curran Associates, Inc.

\bibitem[{Burlot and Yvon(2018)}]{burlot-yvon-2018-using}
Franck Burlot and Fran{\c{c}}ois Yvon. 2018.
\newblock \href {https://doi.org/10.18653/v1/W18-6315} {Using monolingual data
  in neural machine translation: a systematic study}.
\newblock In \emph{Proceedings of the Third Conference on Machine Translation:
  Research Papers}, pages 144--155, Brussels, Belgium. Association for
  Computational Linguistics.

\bibitem[{Clark(1980)}]{clark1980manchu}
Larry Clark. 1980.
\newblock \emph{Manchu suffix list}.
\newblock Department of Asian Languages and Literatures. University of
  Washington.

\bibitem[{Costa-juss{\`a} et~al.(2024)Costa-juss{\`a}, Cross, {\c C}elebi,
  Elbayad, Heafield, Heffernan, Kalbassi, Lam, Licht, Maillard, Sun, Wang,
  Wenzek, Youngblood, Akula, Barrault, Gonzalez, Hansanti, Hoffman, Jarrett,
  Sadagopan, Rowe, Spruit, Tran, Andrews, Ayan, Bhosale, Edunov, Fan, Gao,
  Goswami, Guzm{\'a}n, Koehn, Mourachko, Ropers, Saleem, Schwenk, Wang, and
  Team}]{costa-jussa-etal-2024-scaling}
Marta~R. Costa-juss{\`a}, James Cross, Onur {\c C}elebi, Maha Elbayad, Kenneth
  Heafield, Kevin Heffernan, Elahe Kalbassi, Janice Lam, Daniel Licht, Jean
  Maillard, Anna Sun, Skyler Wang, Guillaume Wenzek, Al~Youngblood, Bapi Akula,
  Loic Barrault, Gabriel~Mejia Gonzalez, Prangthip Hansanti, John Hoffman,
  Semarley Jarrett, Kaushik~Ram Sadagopan, Dirk Rowe, Shannon Spruit, Chau
  Tran, Pierre Andrews, Necip~Fazil Ayan, Shruti Bhosale, Sergey Edunov, Angela
  Fan, Cynthia Gao, Vedanuj Goswami, Francisco Guzm{\'a}n, Philipp Koehn,
  Alexandre Mourachko, Christophe Ropers, Safiyyah Saleem, Holger Schwenk, Jeff
  Wang, and NLLB Team. 2024.
\newblock \href {https://doi.org/10.1038/s41586-024-07335-x} {Scaling neural
  machine translation to 200 languages}.
\newblock \emph{Nature}, 630(8018):841--846.

\bibitem[{Di~Cosmo(2007)}]{di2007diary}
Nicola Di~Cosmo. 2007.
\newblock \emph{The Diary of a Manchu Soldier in Seventeenth-Century China:" My
  Service in the Army", by Dzengseo}.
\newblock Routledge.

\bibitem[{Dubey et~al.(2024)Dubey, Jauhri, Pandey, Kadian, Al-Dahle, Letman,
  Mathur, Schelten, Yang, Fan et~al.}]{grattafiori2024llama3herdmodels}
Abhimanyu Dubey, Abhinav Jauhri, Abhinav Pandey, Abhishek Kadian, Ahmad
  Al-Dahle, Aiesha Letman, Akhil Mathur, Alan Schelten, Amy Yang, Angela Fan,
  et~al. 2024.
\newblock \href {https://arxiv.org/abs/2407.21783} {The llama 3 herd of
  models}.
\newblock \emph{Preprint}, arXiv:2407.21783.

\bibitem[{Edunov et~al.(2018)Edunov, Ott, Auli, and
  Grangier}]{edunov-etal-2018-understanding}
Sergey Edunov, Myle Ott, Michael Auli, and David Grangier. 2018.
\newblock \href {https://doi.org/10.18653/v1/D18-1045} {Understanding
  back-translation at scale}.
\newblock In \emph{Proceedings of the 2018 Conference on Empirical Methods in
  Natural Language Processing}, pages 489--500, Brussels, Belgium. Association
  for Computational Linguistics.

\bibitem[{Elsner and Needle(2023)}]{elsner-needle-2023-translating}
Micha Elsner and Jordan Needle. 2023.
\newblock \href {https://doi.org/10.18653/v1/2023.sigmorphon-1.2} {Translating
  a low-resource language using {GPT}-3 and a human-readable dictionary}.
\newblock In \emph{Proceedings of the 20th SIGMORPHON workshop on Computational
  Research in Phonetics, Phonology, and Morphology}, pages 1--13, Toronto,
  Canada. Association for Computational Linguistics.

\bibitem[{Ghazvininejad et~al.(2023)Ghazvininejad, Gonen, and
  Zettlemoyer}]{ghazvininejad2023dictionarybasedphraselevelpromptinglarge}
Marjan Ghazvininejad, Hila Gonen, and Luke Zettlemoyer. 2023.
\newblock \href {https://arxiv.org/abs/2302.07856} {Dictionary-based
  phrase-level prompting of large language models for machine translation}.
\newblock \emph{Preprint}, arXiv:2302.07856.

\bibitem[{Gorelova(2002)}]{gorelova2002manchu}
Liliya~M Gorelova. 2002.
\newblock Manchu grammar.

\bibitem[{Graham et~al.(2013)Graham, Baldwin, Moffat, and
  Zobel}]{graham-etal-2013-continuous}
Yvette Graham, Timothy Baldwin, Alistair Moffat, and Justin Zobel. 2013.
\newblock \href {https://aclanthology.org/W13-2305/} {Continuous measurement
  scales in human evaluation of machine translation}.
\newblock In \emph{Proceedings of the 7th Linguistic Annotation Workshop and
  Interoperability with Discourse}, pages 33--41, Sofia, Bulgaria. Association
  for Computational Linguistics.

\bibitem[{Haddow et~al.(2022)Haddow, Bawden, Miceli~Barone, Helcl, and
  Birch}]{haddow-etal-2022-survey}
Barry Haddow, Rachel Bawden, Antonio~Valerio Miceli~Barone, Jind{\v{r}}ich
  Helcl, and Alexandra Birch. 2022.
\newblock \href {https://doi.org/10.1162/coli_a_00446} {Survey of low-resource
  machine translation}.
\newblock \emph{Computational Linguistics}, 48(3):673--732.

\bibitem[{Her and Kruschwitz(2024)}]{her-kruschwitz-2024-investigating}
Wan-hua Her and Udo Kruschwitz. 2024.
\newblock \href {https://aclanthology.org/2024.sigul-1.20/} {Investigating
  neural machine translation for low-resource languages: Using {B}avarian as a
  case study}.
\newblock In \emph{Proceedings of the 3rd Annual Meeting of the Special
  Interest Group on Under-resourced Languages @ LREC-COLING 2024}, pages
  155--167, Torino, Italia. ELRA and ICCL.

\bibitem[{Hu(1994)}]{1994}
Zeng~Yi Hu. 1994.
\newblock \href {https://books.google.de/books?id=CCsQAQAAMAAJ} {\emph{A
  comprehensive Manchu-Chinese dictionary}}.

\bibitem[{Hus and Anastasopoulos(2024)}]{hus-anastasopoulos-2024-back}
Jonathan Hus and Antonios Anastasopoulos. 2024.
\newblock \href {https://doi.org/10.18653/v1/2024.emnlp-main.1127} {Back to
  school: Translation using grammar books}.
\newblock In \emph{Proceedings of the 2024 Conference on Empirical Methods in
  Natural Language Processing}, pages 20207--20219, Miami, Florida, USA.
  Association for Computational Linguistics.

\bibitem[{Jacques(2021)}]{Jacques2021}
Guillaume Jacques. 2021.
\newblock \href {https://doi.org/10.5281/zenodo.4548232} {\emph{A grammar of
  {Japhug}}}.
\newblock Number~1 in Comprehensive Grammar Library. Language Science Press,
  Berlin.

\bibitem[{Ko et~al.(2021)Ko, El-Kishky, Renduchintala, Chaudhary, Goyal,
  Guzm{\'a}n, Fung, Koehn, and Diab}]{ko-etal-2021-adapting}
Wei-Jen Ko, Ahmed El-Kishky, Adithya Renduchintala, Vishrav Chaudhary, Naman
  Goyal, Francisco Guzm{\'a}n, Pascale Fung, Philipp Koehn, and Mona Diab.
  2021.
\newblock \href {https://doi.org/10.18653/v1/2021.acl-long.66} {Adapting
  high-resource {NMT} models to translate low-resource related languages
  without parallel data}.
\newblock In \emph{Proceedings of the 59th Annual Meeting of the Association
  for Computational Linguistics and the 11th International Joint Conference on
  Natural Language Processing (Volume 1: Long Papers)}, pages 802--812, Online.
  Association for Computational Linguistics.

\bibitem[{Lample et~al.(2018)Lample, Conneau, Denoyer, and
  Ranzato}]{Lample2018unsupervised}
Guillaume Lample, Alexis Conneau, Ludovic Denoyer, and Marc'Aurelio Ranzato.
  2018.
\newblock \href {https://openreview.net/forum?id=rkYTTf-AZ} {Unsupervised
  machine translation using monolingual corpora only}.
\newblock In \emph{6th International Conference on Learning Representations,
  {ICLR} 2018, Vancouver, BC, Canada, April 30 - May 3, 2018, Conference Track
  Proceedings}. OpenReview.net.

\bibitem[{L{\"a}ubli et~al.(2018)L{\"a}ubli, Sennrich, and
  Volk}]{laubli-etal-2018-machine}
Samuel L{\"a}ubli, Rico Sennrich, and Martin Volk. 2018.
\newblock \href {https://doi.org/10.18653/v1/D18-1512} {Has machine translation
  achieved human parity? a case for document-level evaluation}.
\newblock In \emph{Proceedings of the 2018 Conference on Empirical Methods in
  Natural Language Processing}, pages 4791--4796, Brussels, Belgium.
  Association for Computational Linguistics.

\bibitem[{Le~Scao et~al.(2023)Le~Scao, Fan, Akiki, Pavlick, Ilić, Hesslow,
  Castagné, Luccioni, Yvon, Gallé, Tow, Rush, Biderman, Webson
  et~al.}]{workshop2023bloom176bparameteropenaccessmultilingual}
Teven Le~Scao, Angela Fan, Christopher Akiki, Ellie Pavlick, Suzana Ilić,
  Daniel Hesslow, Roman Castagné, Alexandra~Sasha Luccioni, François Yvon,
  Matthias Gallé, Jonathan Tow, Alexander~M. Rush, Stella Biderman, Albert
  Webson, et~al. 2023.
\newblock \href {https://arxiv.org/abs/2211.05100} {{BLOOM: A 176B-Parameter
  Open-Access Multilingual Language Model}}.
\newblock \emph{Preprint}, arXiv:2211.05100.

\bibitem[{Lin et~al.(2022)Lin, Mihaylov, Artetxe, Wang, Chen, Simig, Ott,
  Goyal, Bhosale, Du, Pasunuru, Shleifer, Koura, Chaudhary, O{'}Horo, Wang,
  Zettlemoyer, Kozareva, Diab, Stoyanov, and Li}]{lin-etal-2022-shot}
Xi~Victoria Lin, Todor Mihaylov, Mikel Artetxe, Tianlu Wang, Shuohui Chen,
  Daniel Simig, Myle Ott, Naman Goyal, Shruti Bhosale, Jingfei Du, Ramakanth
  Pasunuru, Sam Shleifer, Punit~Singh Koura, Vishrav Chaudhary, Brian O{'}Horo,
  Jeff Wang, Luke Zettlemoyer, Zornitsa Kozareva, Mona Diab, Veselin Stoyanov,
  and Xian Li. 2022.
\newblock \href {https://doi.org/10.18653/v1/2022.emnlp-main.616} {Few-shot
  learning with multilingual generative language models}.
\newblock In \emph{Proceedings of the 2022 Conference on Empirical Methods in
  Natural Language Processing}, pages 9019--9052, Abu Dhabi, United Arab
  Emirates. Association for Computational Linguistics.

\bibitem[{Liu et~al.(2024)Liu, Feng, Xue, Wang, Wu, Lu, Zhao, Deng, Zhang, Ruan
  et~al.}]{deepseekai2024deepseekv3technicalreport}
Aixin Liu, Bei Feng, Bing Xue, Bingxuan Wang, Bochao Wu, Chengda Lu, Chenggang
  Zhao, Chengqi Deng, Chenyu Zhang, Chong Ruan, et~al. 2024.
\newblock \href {https://arxiv.org/abs/2412.19437} {Deepseek-v3 technical
  report}.
\newblock \emph{Preprint}, arXiv:2412.19437.

\bibitem[{Marmonier et~al.(2025)Marmonier, Bawden, and
  Sagot}]{marmonier2025explicitlearningllmmachine}
Malik Marmonier, Rachel Bawden, and Benoît Sagot. 2025.
\newblock \href {https://arxiv.org/abs/2503.09454} {Explicit learning and the
  llm in machine translation}.
\newblock \emph{Preprint}, arXiv:2503.09454.

\bibitem[{Merx et~al.(2024)Merx, Mahmudi, Langford, de~Araujo, and
  Vylomova}]{merx-etal-2024-low}
Rapha{\"e}l Merx, Aso Mahmudi, Katrina Langford, Leo~Alberto de~Araujo, and
  Ekaterina Vylomova. 2024.
\newblock \href {https://aclanthology.org/2024.eurali-1.1/} {Low-resource
  machine translation through retrieval-augmented {LLM} prompting: A study on
  the {M}ambai language}.
\newblock In \emph{Proceedings of the 2nd Workshop on Resources and
  Technologies for Indigenous, Endangered and Lesser-resourced Languages in
  Eurasia (EURALI) @ LREC-COLING 2024}, pages 1--11, Torino, Italia. ELRA and
  ICCL.

\bibitem[{Mohammadshahi et~al.(2022)Mohammadshahi, Nikoulina, Berard, Brun,
  Henderson, and Besacier}]{mohammadshahi-etal-2022-small}
Alireza Mohammadshahi, Vassilina Nikoulina, Alexandre Berard, Caroline Brun,
  James Henderson, and Laurent Besacier. 2022.
\newblock \href {https://doi.org/10.18653/v1/2022.emnlp-main.571}
  {{SM}a{LL}-100: Introducing shallow multilingual machine translation model
  for low-resource languages}.
\newblock In \emph{Proceedings of the 2022 Conference on Empirical Methods in
  Natural Language Processing}, pages 8348--8359, Abu Dhabi, United Arab
  Emirates. Association for Computational Linguistics.

\bibitem[{Nordhoff and Hammarstr{\"o}m(2011)}]{nordhoff2011glottolog}
Sebastian Nordhoff and Harald Hammarstr{\"o}m. 2011.
\newblock Glottolog/langdoc: Defining dialects, languages, and language
  families as collections of resources.
\newblock In \emph{First International Workshop on Linked Science 2011-In
  conjunction with the International Semantic Web Conference (ISWC 2011)}.

\bibitem[{Norman(1965)}]{norman1965grammatical}
Jerry Norman. 1965.
\newblock \emph{A grammatical sketch of {Manchu}}.
\newblock University of California Library.

\bibitem[{Norman(2020)}]{norman2020comprehensive}
Jerry Norman. 2020.
\newblock \emph{A comprehensive Manchu-English dictionary}, volume~85.
\newblock BRILL.

\bibitem[{Papineni et~al.(2002)Papineni, Roukos, Ward, and
  Zhu}]{papineni-etal-2002-bleu}
Kishore Papineni, Salim Roukos, Todd Ward, and Wei-Jing Zhu. 2002.
\newblock \href {https://doi.org/10.3115/1073083.1073135} {{B}leu: a method for
  automatic evaluation of machine translation}.
\newblock In \emph{Proceedings of the 40th Annual Meeting of the Association
  for Computational Linguistics}, pages 311--318, Philadelphia, Pennsylvania,
  USA. Association for Computational Linguistics.

\bibitem[{Popovi{\'c}(2015)}]{popovic-2015-chrf}
Maja Popovi{\'c}. 2015.
\newblock \href {https://doi.org/10.18653/v1/W15-3049} {chr{F}: character
  n-gram {F}-score for automatic {MT} evaluation}.
\newblock In \emph{Proceedings of the Tenth Workshop on Statistical Machine
  Translation}, pages 392--395, Lisbon, Portugal. Association for Computational
  Linguistics.

\bibitem[{Post(2018)}]{post-2018-call}
Matt Post. 2018.
\newblock \href {https://doi.org/10.18653/v1/W18-6319} {A call for clarity in
  reporting {BLEU} scores}.
\newblock In \emph{Proceedings of the Third Conference on Machine Translation:
  Research Papers}, pages 186--191, Brussels, Belgium. Association for
  Computational Linguistics.

\bibitem[{Reimers and Gurevych(2019)}]{reimers-2019-sentence-bert}
Nils Reimers and Iryna Gurevych. 2019.
\newblock \href {https://arxiv.org/abs/1908.10084} {Sentence-{BERT}: Sentence
  embeddings using siamese bert-networks}.
\newblock In \emph{Proceedings of the 2019 Conference on Empirical Methods in
  Natural Language Processing}. Association for Computational Linguistics.

\bibitem[{Robertson et~al.(1995)Robertson, Walker, Jones, Hancock-Beaulieu,
  Gatford et~al.}]{robertson1995okapi}
Stephen~E Robertson, Steve Walker, Susan Jones, Micheline~M Hancock-Beaulieu,
  Mike Gatford, et~al. 1995.
\newblock Okapi at {TREC}-3.
\newblock \emph{Nist Special Publication Sp}, 109:109.

\bibitem[{Sennrich et~al.(2016)Sennrich, Haddow, and
  Birch}]{sennrich-etal-2016-improving}
Rico Sennrich, Barry Haddow, and Alexandra Birch. 2016.
\newblock \href {https://doi.org/10.18653/v1/P16-1009} {Improving neural
  machine translation models with monolingual data}.
\newblock In \emph{Proceedings of the 54th Annual Meeting of the Association
  for Computational Linguistics (Volume 1: Long Papers)}, pages 86--96, Berlin,
  Germany. Association for Computational Linguistics.

\bibitem[{Sia et~al.(2024)Sia, DeLucia, and Duh}]{sia-etal-2024-anti}
Suzanna Sia, Alexandra DeLucia, and Kevin Duh. 2024.
\newblock \href {https://doi.org/10.18653/v1/2024.findings-naacl.216}
  {Anti-{LM} decoding for zero-shot in-context machine translation}.
\newblock In \emph{Findings of the Association for Computational Linguistics:
  NAACL 2024}, pages 3403--3420, Mexico City, Mexico. Association for
  Computational Linguistics.

\bibitem[{Song et~al.(2024)Song, Li, Wu, Liu, Lin, and
  Xu}]{song-etal-2024-grammatical}
Huacheng Song, Yi~Li, Yiwen Wu, Yu~Liu, Jingxia Lin, and Hongzhi Xu. 2024.
\newblock \href {https://doi.org/10.18653/v1/2024.wmt-1.117} {How grammatical
  features impact machine translation: A new test suite for {C}hinese-{E}nglish
  {MT} evaluation}.
\newblock In \emph{Proceedings of the Ninth Conference on Machine Translation},
  pages 1200--1221, Miami, Florida, USA. Association for Computational
  Linguistics.

\bibitem[{Tanzer et~al.(2024)Tanzer, Suzgun, Visser, Jurafsky, and
  Melas-Kyriazi}]{tanzer2024a}
Garrett Tanzer, Mirac Suzgun, Eline Visser, Dan Jurafsky, and Luke
  Melas-Kyriazi. 2024.
\newblock \href {https://openreview.net/forum?id=tbVWug9f2h} {A benchmark for
  learning to translate a new language from one grammar book}.
\newblock In \emph{The Twelfth International Conference on Learning
  Representations}.

\bibitem[{Tars et~al.(2022)Tars, Purason, and
  T{\"a}ttar}]{tars-etal-2022-teaching}
Maali Tars, Taido Purason, and Andre T{\"a}ttar. 2022.
\newblock \href {https://aclanthology.org/2022.wmt-1.33/} {Teaching unseen
  low-resource languages to large translation models}.
\newblock In \emph{Proceedings of the Seventh Conference on Machine Translation
  (WMT)}, pages 375--380, Abu Dhabi, United Arab Emirates (Hybrid). Association
  for Computational Linguistics.

\bibitem[{Tiedemann and Scherrer(2017)}]{tiedemann-scherrer-2017-neural}
J{\"o}rg Tiedemann and Yves Scherrer. 2017.
\newblock \href {https://doi.org/10.18653/v1/W17-4811} {Neural machine
  translation with extended context}.
\newblock In \emph{Proceedings of the Third Workshop on Discourse in Machine
  Translation}, pages 82--92, Copenhagen, Denmark. Association for
  Computational Linguistics.

\bibitem[{Touvron et~al.(2023)Touvron, Martin, Stone, Albert, Almahairi,
  Babaei, Bashlykov, Batra, Bhargava, Bhosale, Bikel, Blecher, Ferrer, Chen,
  Cucurull, Esiobu, Fernandes, Fu, Fu, Fuller, Gao, Goswami, Goyal, Hartshorn,
  Hosseini, Hou, Inan, Kardas, Kerkez, Khabsa, Kloumann, Korenev, Koura,
  Lachaux, Lavril, Lee, Liskovich, Lu, Mao, Martinet, Mihaylov, Mishra,
  Molybog, Nie, Poulton, Reizenstein, Rungta, Saladi, Schelten, Silva, Smith,
  Subramanian, Tan, Tang, Taylor, Williams, Kuan, Xu, Yan, Zarov, Zhang, Fan,
  Kambadur, Narang, Rodriguez, Stojnic, Edunov, and
  Scialom}]{touvron2023llama2openfoundation}
Hugo Touvron, Louis Martin, Kevin Stone, Peter Albert, Amjad Almahairi, Yasmine
  Babaei, Nikolay Bashlykov, Soumya Batra, Prajjwal Bhargava, Shruti Bhosale,
  Dan Bikel, Lukas Blecher, Cristian~Canton Ferrer, Moya Chen, Guillem
  Cucurull, David Esiobu, Jude Fernandes, Jeremy Fu, Wenyin Fu, Brian Fuller,
  Cynthia Gao, Vedanuj Goswami, Naman Goyal, Anthony Hartshorn, Saghar
  Hosseini, Rui Hou, Hakan Inan, Marcin Kardas, Viktor Kerkez, Madian Khabsa,
  Isabel Kloumann, Artem Korenev, Punit~Singh Koura, Marie-Anne Lachaux,
  Thibaut Lavril, Jenya Lee, Diana Liskovich, Yinghai Lu, Yuning Mao, Xavier
  Martinet, Todor Mihaylov, Pushkar Mishra, Igor Molybog, Yixin Nie, Andrew
  Poulton, Jeremy Reizenstein, Rashi Rungta, Kalyan Saladi, Alan Schelten, Ruan
  Silva, Eric~Michael Smith, Ranjan Subramanian, Xiaoqing~Ellen Tan, Binh Tang,
  Ross Taylor, Adina Williams, Jian~Xiang Kuan, Puxin Xu, Zheng Yan, Iliyan
  Zarov, Yuchen Zhang, Angela Fan, Melanie Kambadur, Sharan Narang, Aurelien
  Rodriguez, Robert Stojnic, Sergey Edunov, and Thomas Scialom. 2023.
\newblock \href {https://arxiv.org/abs/2307.09288} {Llama 2: Open foundation
  and fine-tuned chat models}.
\newblock \emph{Preprint}, arXiv:2307.09288.

\bibitem[{Vaswani et~al.(2017)Vaswani, Shazeer, Parmar, Uszkoreit, Jones,
  Gomez, Kaiser, and Polosukhin}]{Vaswani2017transformer}
Ashish Vaswani, Noam Shazeer, Niki Parmar, Jakob Uszkoreit, Llion Jones,
  Aidan~N. Gomez, Lukasz Kaiser, and Illia Polosukhin. 2017.
\newblock \href
  {https://proceedings.neurips.cc/paper/2017/hash/3f5ee243547dee91fbd053c1c4a845aa-Abstract.html}
  {Attention is all you need}.
\newblock In \emph{Advances in Neural Information Processing Systems 30: Annual
  Conference on Neural Information Processing Systems 2017, December 4-9, 2017,
  Long Beach, CA, {USA}}, pages 5998--6008.

\bibitem[{Vilar et~al.(2023)Vilar, Freitag, Cherry, Luo, Ratnakar, and
  Foster}]{vilar-etal-2023-prompting}
David Vilar, Markus Freitag, Colin Cherry, Jiaming Luo, Viresh Ratnakar, and
  George Foster. 2023.
\newblock \href {https://doi.org/10.18653/v1/2023.acl-long.859} {Prompting
  {P}a{LM} for translation: Assessing strategies and performance}.
\newblock In \emph{Proceedings of the 61st Annual Meeting of the Association
  for Computational Linguistics (Volume 1: Long Papers)}, pages 15406--15427,
  Toronto, Canada. Association for Computational Linguistics.

\bibitem[{Wei et~al.(2022)Wei, Wang, Schuurmans, Bosma, Ichter, Xia, Chi, Le,
  and Zhou}]{cot2022wei}
Jason Wei, Xuezhi Wang, Dale Schuurmans, Maarten Bosma, Brian Ichter, Fei Xia,
  Ed~H. Chi, Quoc~V. Le, and Denny Zhou. 2022.
\newblock \href
  {http://papers.nips.cc/paper\_files/paper/2022/hash/9d5609613524ecf4f15af0f7b31abca4-Abstract-Conference.html}
  {Chain-of-thought prompting elicits reasoning in large language models}.
\newblock In \emph{Advances in Neural Information Processing Systems 35: Annual
  Conference on Neural Information Processing Systems 2022, NeurIPS 2022, New
  Orleans, LA, USA, November 28 - December 9, 2022}.

\bibitem[{Xue et~al.(2021)Xue, Constant, Roberts, Kale, Al-Rfou, Siddhant,
  Barua, and Raffel}]{xue-etal-2021-mt5}
Linting Xue, Noah Constant, Adam Roberts, Mihir Kale, Rami Al-Rfou, Aditya
  Siddhant, Aditya Barua, and Colin Raffel. 2021.
\newblock \href {https://doi.org/10.18653/v1/2021.naacl-main.41} {m{T}5: A
  massively multilingual pre-trained text-to-text transformer}.
\newblock In \emph{Proceedings of the 2021 Conference of the North American
  Chapter of the Association for Computational Linguistics: Human Language
  Technologies}, pages 483--498, Online. Association for Computational
  Linguistics.

\bibitem[{Yazar et~al.(2023)Yazar, Sahin, and
  Kili{\c{c}}}]{yazar2023lowresource}
Bilge~Kagan Yazar, Durmus~{\"{O}}zkan Sahin, and Erdal Kili{\c{c}}. 2023.
\newblock \href {https://doi.org/10.1109/ACCESS.2023.3336019} {Low-resource
  neural machine translation: {A} systematic literature review}.
\newblock \emph{{IEEE} Access}, 11:131775--131813.

\bibitem[{Yuan et~al.(2024)Yuan, Jiao, Wang, tse Huang, He, Shi, and
  Tu}]{yuan2024gpt4smartsafestealthy}
Youliang Yuan, Wenxiang Jiao, Wenxuan Wang, Jen tse Huang, Pinjia He, Shuming
  Shi, and Zhaopeng Tu. 2024.
\newblock \href {https://arxiv.org/abs/2308.06463} {Gpt-4 is too smart to be
  safe: Stealthy chat with llms via cipher}.
\newblock \emph{Preprint}, arXiv:2308.06463.

\bibitem[{Zhang et~al.(2024{\natexlab{a}})Zhang, Liu, Lin, and
  Feng}]{zhang-etal-2024-teaching}
Chen Zhang, Xiao Liu, Jiuheng Lin, and Yansong Feng. 2024{\natexlab{a}}.
\newblock \href {https://doi.org/10.18653/v1/2024.findings-acl.519} {Teaching
  large language models an unseen language on the fly}.
\newblock In \emph{Findings of the Association for Computational Linguistics:
  ACL 2024}, pages 8783--8800, Bangkok, Thailand. Association for Computational
  Linguistics.

\bibitem[{Zhang et~al.(2024{\natexlab{b}})Zhang, Choi, Song, He, Wang, and
  Li}]{zhang-etal-2024-hire}
Kexun Zhang, Yee Choi, Zhenqiao Song, Taiqi He, William~Yang Wang, and Lei Li.
  2024{\natexlab{b}}.
\newblock \href {https://doi.org/10.18653/v1/2024.findings-acl.925} {Hire a
  linguist!: Learning endangered languages in {LLM}s with in-context linguistic
  descriptions}.
\newblock In \emph{Findings of the Association for Computational Linguistics:
  ACL 2024}, pages 15654--15669, Bangkok, Thailand. Association for
  Computational Linguistics.

\bibitem[{Zoph et~al.(2016)Zoph, Yuret, May, and
  Knight}]{zoph-etal-2016-transfer}
Barret Zoph, Deniz Yuret, Jonathan May, and Kevin Knight. 2016.
\newblock \href {https://doi.org/10.18653/v1/D16-1163} {Transfer learning for
  low-resource neural machine translation}.
\newblock In \emph{Proceedings of the 2016 Conference on Empirical Methods in
  Natural Language Processing}, pages 1568--1575, Austin, Texas. Association
  for Computational Linguistics.

\end{thebibliography}

\appendix

\section{Manchu Language}
\label{sec:appendix A}
Manchu (ISO 639-3: \texttt{mnc}) is a critically endangered Tungusic language native to Northeast China (historically also known as Manchuria). 
Typologically, Manchu is a head-final agglutinative language that makes exclusive use of suffixation for denoting grammatical features, with each suffix having one single function. In this regard, Manchu is often grouped together with other typologically similar languages across Eurasia, e.g., Japanese, Korean, Mongolian, Turkish, and Hungarian.

Manchu is the traditional language of the Manchu people, who founded the Qing dynasty (1644-1911) of China. During this period, Manchu was one of the official government languages, leaving behind a rich collection of historical texts. However, from the 18th century onward, the Manchu language experienced a gradual decline, which accelerated significantly after the fall of the Qing dynasty up until 1980s, at which point the number of native speakers was down to only a few hundreds,\footnote{This number exludes the speakers of the Xibe language (ISO 639-3: \texttt{sjo}), which is sometimes considered as a dialect of Manchu. Xibe is in more vigorous usage and has around 30,000 native speakers. The written form of Xibe is mostly identical to Manchu, aside from several orthographical conventions.}
out of more than 10 million ethnic Manchu population.

Starting from the 1980s, there have been increased efforts to revitalize the Manchu language, as the Manchu communities strive to restore their lost heritage. Primary schools in several Manchu Autonomous Counties, as well as some universities, began offering Manchu courses at various levels. Through the revitalization movements, despite the ultra-small number of genuine native speakers, the number of people who can now speak Manchu as a second language has been steadily growing.

Traditionally, Manchu is written in the Manchu script, an alphabetic writing system that can be easily transliterated into Latin script. All the Manchu data used within our research were already in the form of Latin transliteration.

Manchu is a low-resource language in terms of the available text data on the internet. On the other hand, because of its significant historical importance, Manchu has been extensively studied by generations of linguists and philologists. There exist abundant linguistic resources, including dictionaries, grammar books, some bilingual parallel sentences, and a decent amount of monolingual text, which makes Manchu well-suited for our case study.

\section{Prompts}
\label{sec:appendix prompts}
\begin{itemize}
\item \textbf{Direct Translation:}\\
Please help me translate the following sentence from \textcolor{teal}{\{source language\}} to \textcolor{red}{\{target language\}}:\\
\textcolor{green}{\{sentence\}}\\
Please try your best to translate, it's okay if your translation is bad. Do not refuse to try it. I won't blame you.\\
Please enclose your translation in \#\#\#.
For example, if your translation is ``Hello world'', the last part of your output should be \#\#\# Hello world \#\#\#\\

\item \textbf{Direct Translation with Morphologically Analyzed Sentence:}\\
Please help me translate the following sentence from \textcolor{teal}{\{source language\}} to \textcolor{red}{\{target language\}}:\\
\textcolor{blue}{\{morphologically analyzed sentence\}}\\
The morphemes in this sentence have been segmented: the verb stem and verbal suffixes are separated by '=', the noun stem and nominal suffixes are separated by '$\sim$'. 
Note that some words can be either analyzed as a whole or as a word stem plus a suffix; the different analyses are separated by '/'. In such a case, explanations for both analyses are given below, and you need to choose which one is the most appropriate in the given context.\\
Please try your best to translate, it's okay if your translation is bad. Do not refuse to try it. I won't blame you.\\
Please enclose your translation in \#\#\#.
For example, if your translation is "Hello world", the last part of your output should be \#\#\# Hello world \#\#\#\\

\item \textbf{General Template for Prompts with Components:}\\
Please help me translate the following sentence from \textcolor{teal}{\{source language\}} to \textcolor{red}{\{target language\}}:\\
\textcolor{blue}{\{morphologically analyzed sentence\}}\\
The morphemes in this sentence have been segmented: the verb stem and verbal suffixes are separated by '=', the noun stem and nominal suffixes are separated by '$\sim$'. 
Note that some words can be either analyzed as a whole or as a word stem plus a suffix; the different analyses are separated by '/'. In such a case, explanations for both analyses are given below, and you need to choose which one is the most appropriate in the given context.\\
\textcolor{purple}{\{components\}}

Using all the information provided above, now please translate the sentence into \textcolor{red}{\{target language\}}.
Remember your source sentence is: \textcolor{blue}{\{morphologically analyzed sentence\}}\\
Please enclose your translation in \#\#\#.
For example, if your translation is "Hello world", the last part of your output should be \#\#\# Hello world \#\#\#\\

\item \textbf{Component Dictionary:}\\
For the translation task, you are given the word by word mapping from the \textcolor{teal}{\{source language\}} words to the \textcolor{red}{\{target language\}} words.
Some words can be polysemous and there might be multiple possible English translations. In such a case, please choose the most appropriate one.
Note that for some words, they might be derived from a more basic form, we call this the parent word. The parents are also given in the word-by-word translation.
Here are the dictionary entries for each individual word in the source sentence:\\
\textcolor{orange}{\{dictionary entries\}}

Note that sometimes two or more words can form a collocation and express a specific meaning. You should refer to the collocations listed under the dictionary entries. 
For example, 'mama' means 'grandmother', 'erxe=' means 'to attend', but 'mama erxe=' as a collocation means 'to get smallpox'.
In such a case, explain which collocation meaning you think is most appropriate in the context.

\item \textbf{Component Parallel Examples:}\\
To help with the translation, here are some \textcolor{teal}{\{source language\}}-\textcolor{red}{\{target language\}} parallel sentences that may be helpful for your translation:\\
\textcolor{pink}{\{parallel examples\}}

\item \textbf{Component Grammar:}\\
You are also given this grammar book below. Feel free to rely on this grammar book in your translation task:\\
- Manchu Grammar Book\\
The Manchu language is typologically similar to the Mongolic and Turkic languages. 
All Manchu phrases are head-final; the head-word of a phrase (e.g., the noun of a noun phrase, or the verb of a verb phrase) always falls at the end of the phrase. 
Thus, adjectives and adjectival phrases always precede the noun they modify, and the arguments to the verb always precede the verb. 
As a result, Manchu sentence structure is subject–object–verb (SOV).\\
Manchu also makes extensive use of converb structures and has an inventory of converbial suffixes to indicate the relationship between the subordinate verb and the finite verb that follows it.\\
Unlike English, which uses prepositions, Manchu exclusively uses postpositions.\\
The Manchu language is agglutinative in word structure, meaning that words are formed by adding suffixes to the root, and each morpheme in a word has one distinct meaning or grammatical function.\\
\textcolor{magenta}{\{grammar excerpts\}}"""

\item \textbf{Component Chain-of-Thought Prompting (Annotation):}\\
Given the previous information, please first annotate the meaning and grammatical features of each word in the sentence.\\
For each word, based on their English translation and whether it ends with '='(marker of verb stems), first decide whether the word is nominal (noun/adjective), or a verbal(verb, converb) or else (other part of speech such as adverb, postposition ect.).\\
Then for each noun, please annotate its number (singular/plural) and case (Nominative/Genitive
/Dative-Locative/Accusative/Ablative), based on the particles/suffixes that follow the noun.\\
And for each verb, please annotate its tense (perfect/imperfect) and form (Affirmative/Negative/Interrogative/Imperative/Optative/Desiderative), based on the suffixes attached to the verb.

Then based on the annotations, translate the sentence from \textcolor{teal}{\{source language\}} into \textcolor{red}{\{target language\}} based on the annotations and the analyzed sentence structure. 

\item \textbf{Component Chain-of-Thought Prompting (Annotation + Syntactic Analysis):}\\
Given the previous information, please proceed with the following steps:\\
Step 1:\\
Please first annotate the meaning and grammatical features of each word in the sentence.\\
For each word, based on their English translation and whether it ends with '='(marker of verb stems), first decide whether the word is nominal (noun/adjective), or a verbal (verb, converb) or else (other part of speech such as adverb, postposition etc.).\\
Then for each noun, please annotate its number (singular/plural) and case (Nominative/Genitive
/Dative-Locative/Accusative/Ablative), 
based on the particles/suffixes that follow the noun.\\
And for each verb, please annotate its tense (perfect/imperfect) and form (Affirmative/Negative/Interrogative/Imperative/Optative/Desiderative), based on the suffixes attached to the verb.\\
Step 2:\\
Then based on the annotations, please analyze the sentence structure by figuring out what the subject and object of each verb is. Keep in mind that \textcolor{teal}{\{source language\}}'s basic word order is subject–object–verb (SOV) and it is a head-final language, so that the adjectives and participles always precede the noun they modifies, and the arguments to the verb always precede the verb.\\
Note that clauses can be combined into a single sentence by using converbs, which relate the first action to the second.\\
The final step:\\
Translate the sentence into \textcolor{red}{\{target language\}} based on the annotations and the analyzed sentence structure.

\end{itemize} 

\section{Implementation Details}\label{sec:implementation_details}

\paragraph{Hyperparameters of LLM Generations}
For the Llama3 models, we performed our translation experiments using vLLM.\footnote{\url{https://github.com/vllm-project/vllm}} The model was configured to use half-precision (\texttt{dtype=`float16'}) with a maximum context length of 20,000 tokens. For generations, we used a temperature of 0.9, top‑$p$ sampling with $p = 0.9$, and a maximum output length of 5,000 tokens.

For translation with GPT-4o and DeepSeek-v3, we used OpenAI and DeepSeek's APIs, respectively, with their default settings (i.e., a temperature of 1.0 and top‑$p$ sampling with $p = 1.0$).

\paragraph{Hyperparameters of Fine-Tuning}
For fine-tuning mT5-small, we set the learning rate to 5e-4 and a batch size of 16. We evaluated the model per epoch and employed the early stopping: the training is terminated if no improvement (drop in loss on validation set) is observed over 2 consecutive evaluation steps.
The best-performing checkpoint (with the minimum loss on the validation set) is selected as the final model.

\section{Relative Order of the Parallel Examples and Grammar}\label{sec:order_P_G}
As explained in \secref{Sequential}, the order of adding components is based on which components are expected to be most beneficial, as suggested by previous works. There has been a concern about this order, that the benefit of the parallel sentences from the grammar books might be overshadowed by the retrieved parallel sentences.

To address this concern, we have conducted an additional experiment of comparing $\pi(\mu(\mathbf{x}),\mathrm{D^{*}},\mathrm{G^{*}})$ with the baseline $\pi(\mu(\mathbf{x}),\mathrm{D^{*}})$ to test the benefit of adding Grammar component alone.
Still, the results showed minimal improvement (BLEU: 7.55 → 7.58, chrF: 32.71 → 33.08, SBERT: 61.07 → 59.89), confirming our previous claim in \secref{Grammar} that ``Grammars hardly help''.

\section{Vocabulary Mismatch Between the Test Set and the ICL Data}\label{sec:token-type_overllap}
The vocabulary mismatch between the data used during ICL (dictionary + parallel sentences) and the test set could be a reason for the underestimated BLEU and chrF scores, as mentioned in \ref{performance_underestimated}. In order to measure the similarity between the test set and the data used during ICL, we tokenized all the Manchu and English text involved in the prompts versus the test set, and calculated the number of subword-type overlaps as shown in Table~\ref{tab:token_overlap}:

\begin{table}[h!]
\centering
\begin{tabular}{p{4.2cm}p{1.1cm}p{1.1cm}}
\hline
 & \textbf{English} & \textbf{Manchu}\\
\hline
Subwords in $\mathrm{D}$ and $\mathrm{P}$ & 6,378 & 2,270 \\
Subwords in test set  & 1,353 & 814 \\
Overlapping subwords  & 1,017 & 790 \\
\hline
\% of non-overlapping subwords in test set  & 25\% & 3\% \\
\hline
\end{tabular}
\caption{Subword overlap between the test set and the ICL data.}
\label{tab:token_overlap}
\end{table}

As shown in Table~\ref{tab:token_overlap}, the vocabulary mismatch on the Manchu side is minimal (only 3\%), while the English side exhibits a substantially higher mismatch (25\%). This vocabulary mismatch on the English side could lead to lower BLEU and chrF scores, even when the translation is correct, as is evident in the case of Table~\ref{tab:sbert_scores}. 

\section{Manchu-Chinese Translation}\label{sec:manchu2chinese}
In addition to the Manchu-to-English translation, we have also explored the translation direction of Manchu-to-Chinese. Taking the best setting $\pi(\mu(\mathbf{x}),\mathrm{D^{l+s+c}},\mathrm{P^{bm}})$ from our Manchu-to-English experiments, we use \citet{1994} as the dictionary and the parallel example sentences from \citet{1994} as the parallel corpus. 
We have experimented with GPT-4o and DeepSeek-V3 and the translation is again evaluated using BLEU, chrF, and SBERT. For Chinese word segmentation, we have used Jieba.\footnote{\url{https://github.com/fxsjy/jieba}} The results are shown in Table~\ref{tab:manchu_chinese}:

\begin{table}[h!]
\centering
\begin{tabular}{p{2.5cm}p{1.1cm}p{1.1cm}p{1.1cm}}
\hline
\textbf{Model} & \textbf{BLEU} & \textbf{chrF} & \textbf{SBERT} \\
\hline
GPT-4o & 5.21 & 14.72 & 75.62 \\
DeepSeek-V3 & 9.28 & 19.39 & 79.44 \\
\hline
\end{tabular}
\caption{Manchu-to-Chinese translation performance with GPT-4o and DeepSeek-V3.}
\label{tab:manchu_chinese}
\end{table}

Similar to the trend observed in our Manchu-to-English experiments, DeepSeek-V3 has achieved superior performance compared to GPT-4o, across all three metrics. Interestingly, the SBERT score is noticeably higher for Manchu-to-Chinese translation compared to Manchu-to-English, suggesting that Manchu may be more easily translated into Chinese, likely due to their closer linguistic and cultural affinity. On the other hand, the BLEU and chrF scores are lower for Manchu-to-Chinese. We hypothesize that this is primarily because Chinese MT outputs seem to exhibit lower n-gram overlap with references even when semantically correct, which could be attributed to some inherent characteristics of Chinese, such as more flexible lexical choices and a much larger character inventory. Moreover, recent findings suggest that BLEU and chrF tend to assign disproportionately lower scores to shorter sentences, which may further penalize valid Chinese translations \citep{song-etal-2024-grammatical}.

\onecolumn
\section{Statistical Significance Test}\label{sec:stats_significance}
To test the statistical significance of our results, we have conducted bootstrap resampling with 1000 samples. For each component, the best variant is compared to its baseline to see whether its performance as measured by the BLEU, chrF, and SBERT scores, is statistically significantly higher than the baseline. 

\begin{table*}[h!]
	\centering
	\setlength{\tabcolsep}{6pt}
	\begin{tabular}{llccc}
		\toprule
		\textbf{Hypothesis tested (variant > baseline)} & \textbf{Metric} & \textbf{Variant} & \textbf{Baseline} & \textbf{\textit{p}-value} \\
		\midrule
		
		\multirow{3}{*}{$\pi(\mu(x), \mathbf{D_{lsc}}) > \pi(x)$} 
		& BLEU   & 7.55*** & 3.44 & 0.0009 \\
		& chrF   & 32.71*** & 21.86 & 0.0009 \\
		& SBERT  & 61.07*** & 34.21 & 0.0009 \\
		
		\midrule
		
		\multirow{3}{*}{$\pi(\mu(x), \mathrm{D_{lsc}}, \mathbf{P_{bm}}) > \pi(\mu(x), \mathrm{D_{lsc}})$}
		& BLEU   & 8.84*  & 7.55 & 0.04 \\
		& chrF   & 33.72* & 32.71 & 0.03 \\
		& SBERT  & 61.35 & 61.07 & 0.3 \\
		
		\midrule
		
		\multirow{3}{*}{$\pi(\mu(x), \mathrm{D_{lsc}}, \mathrm{P_{bm}}, \mathbf{G_{lp}}) > \pi(\mu(x), \mathrm{D_{lsc}}, \mathrm{P_{bm}})$}
		& BLEU   & 8.90 & 8.84 & 0.4 \\
		& chrF   & 33.77  & 33.72 & 0.4 \\
		& SBERT  & 60.40 & 61.35 & 0.9 \\
		
		\midrule
		
		\multirow{3}{*}{$\pi(\mu(x), \mathrm{D_{lsc}}, \mathrm{P_{bm}}, \mathrm{G_{lp}}, \mathbf{C_{as}}) > \pi(\mu(x), \mathrm{D_{lsc}}, \mathrm{P_{bm}}, \mathrm{G_{lp}})$}
		& BLEU   & 8.49  & 8.90 & 0.4 \\
		& chrF   & 33.43 & 33.77 & 0.7 \\
		& SBERT  & 59.01 & 60.40 & 0.9 \\
		
		\bottomrule
	\end{tabular}
	\caption{Statistical significance (p-values) of BLEU, chrF and SBERT scores for each added component over its immediate baseline. Asterisks denote statistical significance: *\textit{p} < 0.05, **\textit{p} < 0.01, ***\textit{p} < 0.001.}
	\label{tab:component_significance}
\end{table*}
As shown in Table~\ref{tab:component_significance}, the performance gains of adding the best variant of Dictionary and Parallel Examples are statistically significant (with the only exception of SBERT for Parallel Examples), while for Grammar and CoT the performance is not statistically significantly better than the baseline, which aligns with our claims. In the case of Parallel Examples, although the p-value for SBERT is not statistically significant, both BLEU and chrF show significant improvements. Therefore, considering all three metrics together, we still find it valid to claim that Parallel Examples is beneficial.

\section{Instructions Given to Human Raters}\label{sec:appendix instructions}
\begin{tcolorbox}[breakable]
In this evaluation task, you will evaluate 99 English translations of Manchu sentences.
For each evaluation item, you will be shown:
\begin{itemize}
    \item The original Manchu sentence
    \item An English reference sentence
    \item A system-generated English translation
\end{itemize}
Your evaluation should be based on how well the translation captures the meaning of the original Manchu sentence. The English reference translation is included only to help resolve ambiguities — do not score based on how closely the system-generated translation matches the reference. If the system-generated translation uses different words or phrasing but adequately conveys the meaning, it should not be penalized.

The focus of this evaluation is on how well the system-generated English translation conveys the meaning of the original Manchu sentence. Do not penalize translations for awkward or unnatural English phrasing as long as the meaning is adequately preserved.

You will use a slider (0\%–100\%) to score each translation. Below are example cases to help guide your judgments:
\begin{itemize}
\item 0\% – No meaning preserved

Manchu: miyoose tusy labdu tacin umesi oshon ehe

Reference: There are many chieftains of the Miao, whose customs are cruel and wicked in the extreme.

Translation: The third month after the harvest is very difficult.

→ Almost none of the source meaning is captured. An extremely low score is appropriate.

\item 33\% – Some meaning preserved

Manchu: bayara be beise gaifi morin ulebumbi

Reference: The Beise led the bayara guards to feed the horses.

Translation: The prince, taking joy, rides the horse.

→ A few words are correctly translated (e.g., "beise", "morin"), but the overall meaning is not.

\item 66\% – Most meaning preserved

Manchu: emu inenggi ududu morin gabtabumbi

Reference: In a single day many horses were shot.

Translation: One day, several horses shone forth.

→ Most of the meaning is preserved, but there is a key error ("gabtabumbi" mistranslated).

\item 100\% – Adequate translation, all meaning preserved

Manchu: bi gemun hecen i baru genembi

Reference: I'm going to Beijing (the capital).

Translation: I go toward the capital city.

→ The system output preserves the full meaning, even if the wording differs from the reference.
\end{itemize}
Please try to be consistent in your use of the scale across all items. Your valuable evaluation will help improve the quality of machine translation for endangered languages like Manchu.
\end{tcolorbox}

\section{Output Examples}\label{sec:output_examples}

\begin{table*}[h!]
\centering
\begin{tabular}{p{2.1cm}p{7.5cm}p{4.4cm}}
\toprule
    & Retrieved Dictionary Entries\newline
    (Input Sentence: \textit{se udu oho})  & Translation \\ 
\midrule
$\pi(\mu(\mathbf{x}))$ & \textbf{morphological segmentation:}\newline
 se udu oho/o-ho.\newline
 \textsl{(alternative analyses are separated by `/' )} & It happened to be the time. \\
$+\mathrm{D^{l}}$ & \textbf{+lexical entries:}\newline
 se: {\color{blue}1. year (said of age), age} 2. raw silk, unprocessed silk 3. the juncture of the stem and root on the ginseng plant 4. (plural suffix)\newline
 udu: {\color{orange}1. ``How many?", ``How much?"} 2. several 3. although \newline
oho: armpit \newline
o-: 1. to become, to change into 2. {\color{red}to be, to exist} 3. to be proper, to be permissible
& {\color{orange}\underline{How many}} {\color{red}\underline{are}} the {\color{blue}\underline{years}}? \\
 \midrule
Ground Truth & & {\color{orange}\underline{How many}} {\color{blue}\underline{years}} old? \\
\bottomrule
\end{tabular}
\caption{Comparison between the outputs of $\pi(\mu(\mathbf{x}))$ and $\mathrm{D^{l}}$. The LLM selects the most appropriate sense, interpreting \textit{se} as ``year'' and \textit{udu} as ``How many''. Additionally, the LLM correctly identifies that \textit{oho} is better analyzed as \textit{o-ho}, where \textit{o-} means ``to be'', rather than as \textit{oho} (``armpit'') which does not fit in this context.}
\label{example:disambiguation}
\end{table*}

\begin{table*}[h!]
\centering
\begin{tabular}{p{2.1cm}p{7.5cm}p{5cm}}
\toprule
    & Retrieved Dictionary Entries\newline
    (Input Sentence: \textit{gvsai ejen jiha...})  & Translation \\ 
\midrule
$\mathrm{D^{l}}$ & \textbf{lexical entries:}\newline
 gvsa: banner \hspace{1cm} ejen: master, ruler\newline
 ji-: to come & the ruler of the banner comes...\\
$\mathrm{D^{l+s}}$ & \textbf{+suffixes:}\newline
 -i: it is used to indicate possession (Genitive) or the means by which something is accomplished (Instrumental)\newline
 {\color{orange}-ha: Perfect participle: ara- 'to do', ara-ha 'he.did, who has done, what was done'} & the ruler of the banner {\color{orange}\underline{came}}... \\
$\mathrm{D^{l+s+c}}$ & \textbf{+collocations:}\newline
 {\color{blue}gvsa-i ejen: Lieutenant-General (of a banner)} & the {\color{blue}\underline{Lieutenant-General}} {\color{orange}\underline{came}}... \\ \midrule
Ground Truth     &      & Lieutenant-general came...  \\ 
\bottomrule         
\end{tabular}
\caption{Comparison of the outputs for $\mathrm{D^{l}}$, $\mathrm{D^{l+s}}$ and $\mathrm{D^{l+s+c}}$. As the example shows, the additional information about suffixes and collocations helps the model to correctly translate the tense of the verb and the phrase \textit{gvsai ejen}.}
\label{tab:output_dict}
\end{table*}

\begin{table*}[h!]
\centering
\begin{tabular}{p{3.2cm}p{6cm}p{5.5cm}}
\toprule
    & Retrieved Parallel Examples\newline
    (Input Sentence: \textit{geli emu hvlha be we waha be sarkv.})  & Translation \\ 
\midrule
$\pi(\mu(\mathbf{x}), \mathrm{D^{l+s+c}})$ & - & Still, \underline{it is unknown} who has killed one bandit.\\
$+\mathrm{P^{r}}$ & ... & Also, one bandit, who has killed, \underline{is not known}.\\
$+\mathrm{P^{d}}$ & ... & Also, who has killed one bandit \underline{without knowing it}?\\
$+\mathrm{P^{bm}}$ & Manchu: wede hvlhame gamabuha be inu sarkv.\newline
English: {\color{orange}I} don't know who stole it.\newline... & Still, \underline{{\color{orange}I} do not know} who killed one bandit. \\ \midrule
Ground Truth     &      & Who killed the other rebel, \underline{{\color{orange}I} do not know}.  \\ \bottomrule
\end{tabular}
\caption{Comparison between the outputs of the baseline $\pi(\mu(\mathbf{x}), \mathrm{D^{l+s+c}})$ and the variants of adding parallel examples retrieved by different ways ( $+\mathrm{P^{r}}$, $+\mathrm{P^{d}}$ and $+\mathrm{P^{bm}}$). As the example demonstrates, when using the BM25 algorithm, the retrieved parallel example helps the model to recover the subject `I', which is often omitted in Manchu but can be inferred from the context.}
\label{example:output_para}
\end{table*}

\begin{table*}[h!]
\centering
\begin{tabular}{p{3.2cm}p{8cm}p{2.5cm}}
\toprule
    & Retrieved Grammar Excerpts\newline
    (Input Sentence: \textit{muke \underline{be} genekini})  & Translation \\ 
\midrule
$\pi(\mu(\mathbf{x}),\mathrm{D^{l+s+c}},\mathrm{P^{bm}})$ & \textbf{lexical entries:}\newline
be: accusative particle & go {\color{blue}\underline{by}} water\\
$+\mathrm{G^{s}}$ & \textbf{+grammar excerpts:}\newline
  An object of a verb having definite or specific reference is shown with the particle be.
& go \underline{to} the water \\
$+\mathrm{G^{l}}$ & \textbf{+grammar excerpts:}\newline
...{\color{blue}The accusative may express space within and means of conveyance by which the motion is going on.} & go {\color{blue}\underline{by}} water\\
$+\mathrm{G^{l+p}}$ & \textbf{+grammar excerpts +example:}\newline 
...{\color{blue}The accusative may express space within and means of conveyance by which the motion is going on.}\newline
  jugvn {\color{blue}be} yabu-me\newline
  road {\color{blue}ACC} go-CONV “to go {\color{blue}along} the road;”
& go {\color{blue}\underline{by}} water\\
\midrule
Ground Truth     &      & travel {\color{blue}\underline{by}} water  \\ \bottomrule
\end{tabular}
\caption{Comparison between the outputs of the baseline $\pi(\mu(\mathbf{x}),\mathrm{D^{l+s+c}},\mathrm{P^{bm}})$ and the variants of adding grammar excerpts( $+\mathrm{G^{l}}$, $+\mathrm{G^{l+p}}$ and $+\mathrm{G^{s+p}}$). With the more detailed long grammar excerpt explaining a specific function of Manchu accusative as \textit{expressing space within and means of conveyance by which the motion is going on}, the model correctly selects the English preposition `by'. In contrast, when provided with the short grammar excerpt of $\mathrm{G^{s+p}}$, the model chooses the preposition `to' as the translation for the accusative particle. This likely stems from the model's broader knowledge of the accusative case, including its usage in other languages such as German, where it can correspond to the English preposition `to'. However, in the case of Manchu, such a translation is incorrect. Notably, in the baseline $\pi(\mu(\mathbf{x}),\mathrm{D^{l+s+c}},\mathrm{P^{bm}})$ without any grammar excerpts, the model also translates the preposition correctly, either by chance or through contextual inference.}
\label{example:grammar}
\end{table*}

\begin{table*}[h!]
\centering
\begin{tabular}{p{4cm}p{4cm}p{6cm}}
\toprule
    & Translation\newline
    (Input Sentence: \textit{nikan coohiyan tob seme emu dulin})  & Model Response \\ 
\midrule
$\pi(\mu(\mathbf{x}),\mathrm{D^{l+s+c}},\mathrm{P^{bm}},\mathrm{G^{l+p}})$ & Chinese and Korea are exactly in one middle & 
... \\
$+\mathrm{C^{a}}$ 
 & {\color{red}Halfway} between China and Korea & Now, let's analyze ... "emu dulin" modifies the action, indicating the manner or status as "one middle" or "a middle stage" contextually suggesting {\color{red}halfway} or middle grounds ...\\
$+\mathrm{C^{a+s}}$ 
& {\color{red}The statement or identity concerning} the Chinese and Korea is exactly {\color{red}as described} in the middle. & Step 2: Analyze the sentence structure ... 
The sentence implies a direct {\color{red}identification or explanation regarding} the entities "Chinese" (nikan) and "Korea" (coohiyan) with respect to something being exactly in the middle or balanced ("emu dulin") {\color{red}described} as "just" or "right" ...\\ \midrule
Ground Truth    & Chinese and Korean each just one half.  & \\ \bottomrule
\end{tabular}
\caption{The translations and some of the model responses involving CoT steps, comparing the baseline with the variants of adding CoT prompting instruction($+\mathrm{C^{a}}$ and $+\mathrm{C^{a+s}}$). As shown in the model responses, the CoT prompting makes the model engage in multiple iterative reasoning steps. While these steps are intended to help the translation, it may also introduce errors at each turn, leading to a worse final translation.}
\label{example:cot_unhelpful}
\end{table*}

\begin{table*}[h!]
\centering
\begin{tabular}{p{1.1cm}p{8.5cm}p{5cm}}
\toprule
    & Retrieved Dictionary Entries\newline
    (Input Sentence: \textit{mini morisa be ganabufi...})  & Translation \\ 
\midrule
$\mathrm{D^{l}}$ & \textbf{lexical entries:}\newline
 mini: my, of me \hspace{1cm} morin: horse\newline
 be: 1. we (exclusive) 2. (accusative particle)
ganabu-: (causative of gana-); gana-: to fetch
 & After my \underline{horse{\color{orange}s}} were fetched...\\
 $\mathrm{D^{l}_{e}}$ & \textbf{lexical entries:}\newline (\textbf{enciphered sentence:} nopo nusote ci hepecago)\newline
 nopo: my, of me \hspace{1cm} nusop: horse\newline
 ci: 1. we (exclusive) 2. (accusative particle)
hepeca-: (causative of hepe-); hepe-: to fetch
 & My \underline{horse} was caused to fetch...\\
$\mathrm{D^{l+s}}$ & \textbf{+suffixes:}\newline
 -fi: Perfect converb: ara- 'to do', ara-fi 'having done, he did and (then did something else)'.\newline
 {\color{orange}-sa: (-sa/-se/-so) Plural: sakda 'old man', sakda+sa 'old men'.} & My \underline{horse{\color{orange}s}} having been fetched...\\ \midrule
Ground Truth &      & I had my \underline{horse{\color{orange}s}} fetched and... \\ \bottomrule         
\end{tabular}
\caption{Comparison of the outputs betweem $\mathrm{D^{l}}$, $\mathrm{D^{l}_{e}}$ and $\mathrm{D^{l+s}}$. This example suggests that the LLM already has some prior knowledge about the Manchu plural suffix so that even when the information about plural suffix \textit{-sa} is not included in the prompt of $\mathrm{D^{l}}$, the model is still able to identify the plurality. In contrast, when the Manchu tokens are enciphered, there is no clue for plurality in the context, leading the model to incorrectly identify the noun as singular.}
\label{tab:encrypt}
\end{table*}

\begin{table*}[h!]
\centering
\begin{tabular}{p{2.4cm}p{7cm}p{1cm}p{1cm}p{1cm}}
\toprule
    Model & Translation \newline(Input Sentence: \textit{ere uthai tere gucu inu})  & BLEU & chrF& SBERT\\ 
\midrule
Llama3-1B & te=re/tere inu uju be tongki & 0.0 & 7.75 & 1.67\\
Llama3-3B & This friend and that one are the same, at the same time, or both are friends. & 2.86 & 27.46 & 42.74 \\
Llama3-8B & Even though he/she/it sits, a friend. & 6.77 & 31.36 & 42.81
\\
Llama3-70B & This one, then, is also living with a friend. & 8.05 & 35.86 & 46.97 \\ 
GPT-4o & This immediately is that friend too & 19.3 & \textbf{45.72} & 70.88 \\
DeepSeek-V3 & This is also that friend & \textbf{20.8} & 45.26 & \textbf{81.96}\\
\midrule
Ground Truth     &   This is that very friend.  &\\ \bottomrule
\end{tabular}
\caption{Comparison of outputs across different LLMs, along with their \textbf{sentence-level} BLEU and chrF scores, as well as SBERT scores. We \textbf{bold} the best score for each metric. In this example, Llama3-1B fails to follow the instructions, generating a sentence in Manchu. This illustrates a clear trend that as models become larger and/or more advanced, the translation quality consistently improves.}
\label{example:size}
\end{table*}

\end{document}